\documentclass{article}

\usepackage{arxiv}

\usepackage[utf8]{inputenc} 
\usepackage[T1]{fontenc}    
\usepackage{hyperref}       
\usepackage{url}            
\usepackage{booktabs}       
\usepackage{amsfonts}       
\usepackage{nicefrac}       
\usepackage{microtype}      
\usepackage{lipsum}

\usepackage{graphicx}
\usepackage{amsmath}
{}
{}
{}

\usepackage{multirow}
\usepackage{rotating}
\usepackage[table]{xcolor}
\usepackage[finalnew]{trackchanges}

\title{Unknown Presentation Attack Detection against Rational Attackers}

\author{
  Ali Khodabakhsh\thanks{\href{http://ali.khodabakhsh.org}{http://ali.khodabakhsh.org}} \\
  Department of Information Security and Communication Technology\\
  Norwegian University of Science and Technology\\
  2815 Gj\o vik, Norway \\
  \texttt{ali.khodabakhsh@gmail.com}
\And
  Zahid Akhtar \\
  Department of Network and Computer Security\\
  State University of New York Polytechnic Institute\\
  13502 Utica, USA \\
  \texttt{akhtarz@sunypoly.edu}
}

\begin{document}
\maketitle

\begin{abstract}
Despite the impressive progress in the field of presentation attack detection and multimedia forensics over the last decade, these systems are still vulnerable to attacks in real-life settings. Some of the challenges for existing solutions are the detection of unknown attacks, the ability to perform in adversarial settings, few-shot learning, and explainability. In this study, these limitations are approached by reliance on a game-theoretic view for modeling the interactions between the attacker and the detector. Consequently, a new optimization criterion is proposed and a set of requirements are defined for improving the performance of these systems in real-life settings. Furthermore, a novel detection technique is proposed using generator-based feature sets that are not biased towards any specific attack species. To further optimize the performance on known attacks, a new loss function coined categorical margin maximization loss (C-marmax) is proposed which gradually improves the performance against the most powerful attack.  The proposed approach provides a more balanced performance across known and unknown attacks and achieves state-of-the-art performance in known and unknown attack detection cases against rational attackers. Lastly, the few-shot learning potential of the proposed approach is studied as well as its ability to provide pixel-level explainability.
\end{abstract}


\section{Introduction}

Over the last decades, there have been major break-throughs in the fields of manufacturing, computing, and communication, resulting in cost reduction as well as higher availability of manufacturing and synthesis processes to the public. 
Among the beneficiaries of these advancements are the attackers to biometric and forensic systems, who take advantage of these methods to devise new and more powerful attacks. 
Relying on the fact that face is the main modality of human communication in daily life, and the ever-growing interest in the use of face biometrics in real-life applications, methods that can realistically produce facial videos have an immense potential for abuse. The infamous Deepfake tool\footnote{\href{https://github.com/deepfakes/faceswap}{https://github.com/deepfakes/faceswap}} is such an example that has repeatedly been used for the purpose of fake news generation to such an extent that a bill was passed in the US senate to report at specified intervals on the state of digital content forgery technology.

Consequently, biometric and forensic systems face new challenges every day as they have to become secure against a wider range of attacks happening at a higher frequency.
Making the matters worse, the existing detection solutions are often designed against a specific attack (or set of attacks) in controlled environments and lack the capacity to face the challenges of real-life deployment.
This is evident from the results of the recent Deepfake detection challenge\footnote{\href{https://www.kaggle.com/c/deepfake-detection-challenge}{https://www.kaggle.com/c/deepfake-detection-challenge}} organized by Facebook where the best performing algorithm had a detection rate of only $65\%$ when faced with unknown generation techniques.
As such, addressing vulnerabilities of existing solutions and introduction of methods to mitigate these vulnerabilities is of utmost importance for deployment of these systems in practice.

One aspect of challenges for deployment that is rarely studied is the selection process of a rational attacker. 
It is expected for an attacker with an ever-growing menu of options for attacking to behave rationally and choose the most powerful attack available to him to maximize the chance of infiltration. 
Furthermore, as the defender does not have knowledge or access to massive amounts of data for all possible attacks available to attackers, his detector would probably be tasked with the detection of unknown attacks or attacks from which only a few training examples are available.
Additionally, lack of explainability limits the use of a system in high-stake applications where explainability increases its utility when operated by a human supervisor.

In this article, to address these challenges, a game-theoretic approach is considered for the formulation of the interactions between the attacker and the detector.
Resulting from this, an optimization criterion is formulated and a set of requirements are defined for designing the detector accordingly.
To tackle the problem of unknown attack detection and few-shot learning, the use of unbiased compressed feature sets is proposed, and for targeting the optimal performance, a new loss function is defined faithful to the formulated optimization criteria. Finally, the explainability of the proposed method is demonstrated with a few examples. 
The rest of this article is organized as follows:
In section \ref{litreview}, the related literature is reviewed and a theoretic basis for the proposed approach is established in section \ref{theory}. Afterward, the proposed method is introduced in section \ref{method} and the case study experiment setup is explained in section \ref{setup}. Finally, the results of the experiments are reported and analyzed in section \ref{pad} and \ref{deepfake} and the article is concluded in section \ref{conclusion}.

\section{Literature Review}
\label{litreview}
Considering the task of forgery detection or presentation attack detection on the face modality, there exist three relevant threads of research. First is the field of multimedia forensics, and more specifically, anti- counter forensics (CF). This thread of research takes an adversarial view on the problem and tries to optimize the performance of the detection system facing an adversary who is actively working towards undermining the performance of the detector. Second is the field of presentation attack detection (PAD) in which the objective of the detector is to secure a biometric system against attacks from different presentation attack species (PAS). Lastly, the newly established thread of Deepfake detection is considered that was initiated to address new phenomenon of availability of automated open-source photo-realistic digital video manipulation techniques on the internet. In this article, the terminology proposed for the field of presentation attack detection is relied on. Consequently, the act of forgery is called attacking the detector and generation techniques used by the forger are called attack species.

\subsection{Anti Counter Forensics}
The majority of solutions in the literature are designed neglecting the fact that an attacker works actively to undermine the performance of the detection system \cite{Bohme2013}.
To address the vulnerability to CF attacks, many anti-CF techniques have been developed, with a focus on detecting the traces left by CF techniques.
Anti-forensic techniques often target a specific CF technique, and as a result, an obvious problem occurs when the attacker anticipates the use of the anti-CF technique and adjust accordingly.
In turn, the defender would need to resort to the introduction of a new detection system to detect the anti-CF attacks, resulting in a never-ending iterative loop with unforeseeable outcomes \cite{6515027}. A possible solution to this problem is to design techniques that are intrinsically more resistant to CF attempts \cite{4761645}\cite{10.1145/2037252.2037256}.
For example, in \cite{7004806} the authors proposed the use of second-order statistics derived from co-occurrence matrices and show robustness against CF attacks. 
Zhang et al. \cite{7090993}\change{}{ used a reduced feature set based on assumptions on the attacker's data manipulation strategy.}
A combination of one-class and two-class classifiers is proposed in \cite{10.1007/978-3-319-20248-8_15}. 
Another interesting approach is the randomization of the feature selection process \cite{8653427}.
In \cite{7823902} and \cite{8081213}, the authors propose the reuse of the original feature space for the detection of CF attacks by retraining for the task of double JPEG compression detection. The third group of solutions rely on game theory to model the interactions between the detector and the attacker and improve the performance of the detector at the final equilibrium \cite{8553305} \cite{6288237}\cite{6400246}.
All aforementioned methods address the case where attacker has a limited choice of CF attacks and do not consider  selection process of attacks in optimization of the detector.

\subsection{Presentation Attack Detection}
Similar to anti-CF techniques, the existing PAD research can be categorized into three branches: (1) PAD systems that address specific PASs, (2) PAD systems that increase or optimize the feature set to detect a higher variety of attacks, and finally, (3) PAD systems that rely on game theory to model the interactions between attacker and defender and optimize the PAD performance accordingly.

The early PAD methods addressed PAD for specific PAS, examples of which are methods relying on features such as blinking, head movement, and textures \cite{4409068, 4291551, 10.1007/978-3-319-46654-5_67}.
\change{Features such as 2D Fourier spectrum }{}\change{, local binary patterns (LBP) }{}\change{, histogram of oriented gradients }{}\change{, difference of Gaussians }{}\change{, scale invariant feature transform }{}\change{, and speeded up robust features }{}\change{ have been used for print and replay attack detection. In }{}\change{, the authors propose the use of central difference convolutional networks which use central deference convolutions that are designed after LBP features }{}\change{. In }{}\change{ the authors investigate the use of layer-by-layer progressive compact space generation for achieving large inter-class and small intra-class distances. Generation of additional training samples with style transfer has also been studied to create user-centered detection models in }{}\change{.}{
Different Features have been used }\cite{6612955}, \cite{10.1007/978-3-642-15567-3_37}, \cite{7487030},\cite{7748511}\change{}{, such as 2D Fourier spectrum }\cite{7031384}\change{}{, local binary patterns (LBP) }\cite{10.1007/978-3-642-37410-4_11}\change{}{. Authors in }\cite{Yu_2020_CVPR}\change{}{, }\cite{Li2020compactnet}\change{}{ and }\cite{israel2019style}\change{}{ presented central difference convolutional networks, layer-by-layer progressive compact space generation, and style transfer techniques, respectively.} Many PAD methods rely on an augmented feature set using additional hardware. Examples include 3D depth camera \cite{WANG2017332}, multi-spectral camera \cite{5771438}, and microphones \cite{5584864}. 
However, these techniques require the addition of often expensive hardware to the pipeline, which may not be feasible in all applications. \change{A few studies try to use generalizable feature sets for PAD. One feature set that is considered for this purpose is image-quality related features.
In }{}\change{, the authors propose the use of image distortion analysis using specular reflection, blurriness, chromatic moment, and color diversity as features. }{
A few studies tried to use generalizable feature sets for PAD. In }\cite{7031384}\change{}{, the authors propose the use of image distortion analysis.}
The use of $25$ general image quality features for PAD is investigated in \cite{6671991}.
In \cite{7821027}, a regression function is learned to map the image quality assessment scores. The use of pixel-level supervision for improving features is investigated in \cite{Liu_2018_CVPR} and regional self-supervision in \cite{deb2020look}. A limited number of studies tried to address the generalizability of PAD systems \cite{ramachandra2017presentation} \cite{Bhattacharjee2019recent} using a one-class classification approach \cite{7984788} \cite{8411206}, deep metric learning model \cite{Perez-Cabo_2019_CVPR_Workshops}, and zero-shot \cite{Liu_2019_CVPR}. To the best of our knowledge, no game-theoretic approach is proposed to model interactions between  attacker and  defender.

\subsection{DeepFakes Detection}
\change{
The approaches for Deepfake detection can be categorized into similar categories. 
Among proposed approaches that try to detect samples from a specific generation technique, a detection approach is proposed in}{}
\change{for computer-generated imagery (CGI) based on a lack of asymmetry in generated images. 
The spatiotemporal deformations of a 3D face model are also explored for the detection of CGI in}{} 
\change{relying on the fact that natural faces show a wider range of deformation compared to the synthetic ones. 
The use of periodic blood flow is considered as a discriminative feature in}{}
\change{. Similar methods have been applied for the detection of Deepfakes relying on generation flaws}{} 
\change{, blinking}{}
\change{, and blood flow }{}
\change{Face warping artifacts have been studied in}{}
\change{ relying on the resolution difference in the spliced portion of the video.
Further studies have investigated the use of face landmark locations}{}
\change{and head pose consistencies}{}
\change{.
Generative adversarial networks (GAN) produce certain artifacts relating to color synthesis and saturation which are used for detection in}{} 
\change{and }{}
\change{respectively. 
The idea of detecting GAN-generated images using architecture-specific GAN fingerprints was proposed in}{} 
\change{A number of more general-purpose detection methods are introduced recently, including the use of mesoscopic features }{}\change{,general-purpose deep convolutional neural networks (CNN)}{}\change{, attention mechanism}{}\change{and capsule networks}{}\change{.Some methods take into account the temporal aspect of videos and propose the use of long short-term memory (LSTM) networks}{}\change{, recurrent CNNs}{}\change{, and optical fields }{}\change{.}{}
\change{A number of more general-purpose detection methods have been introduced recently, including use of mesoscopic features }{}
\change{, attention mechanism }{}
\change{and capsule networks}{}
\change{. Some methods analyzed temporal aspect of videos via long short-term memory (LSTM) networks }{}
\change{, recurrent CNNs}{}
\change{, and optical fields }{}
\change{. It has been shown that most of these detectors tend to overfit to the known attacks and show limited generalizability }{}
\change{.The problem of generalization has been studied in a few articles. An auto-encoder based detection schemes were proposed in}{}
\change{and}{} 
\change{Incremental learning has also been proposed for adaptation to new attacks}{}
\change{To avoid the detector from focusing on low-level GAN artifacts, these artifacts are suppressed in a preprocessing step in}{}
\change{, while in}{} 
\change{transferability of the network is improved by pre- and post-processing.}{}
\change{The time dimension has also been utilized in 
with attention mechanism.}{}
\change{Other works can be seen in}{}
\change{Most studies have a heavy focus on GAN generated images and do not consider other types of manipulations such as CGI based methods and face-swapping.
Furthermore, none of the aforementioned studies take into account the rationality of the attacker nor the case in which the attacker has multiple choices of attack species. A summary of the most representative works in anti counter forensics, presentation attack detection and deepfakes detection is presented in Table \ref{tabel1:summary}.}{Several approaches have been proposed for detecting DeepFakes, such as lack of asymmetry in computer-generated imagery}\cite{6333919}, spatio-temporal deformations of a 3D face model \cite{7097661}, use of periodic blood flow \cite{7025049}, generation flaws \cite{8638330}, blinking \cite{8630787}, and blood flow \cite{9141516}, face warping artifacts \cite{Li_2019_CVPR_Workshops}, use of face landmark locations \cite{10.1145/3335203.3335724}, head pose consistencies \cite{8683164}, mesoscopic features \cite{8630761}, architecture-specific GAN fingerprints \cite{DBLP:journals/corr/abs-1808-07276} \cite{8803661} \cite{8695364}, convolutional neural networks (CNNs) \cite{Rossler_2019_ICCV}, attention mechanism \cite{Dang_2020_CVPR} and capsule networks\cite{8682602}, long short-term memory (LSTM) networks \cite{8639163}, recurrent CNNs \cite{Sabir_2019_CVPR_Workshops}, and optical fields \cite{Amerini_2019_ICCV}. However, such detectors tend to overfit to the known attacks and show limited generalizability \cite{8553251}. The problem of generalization has been studied in a few articles using, e.g., auto-encoder in \cite{DBLP:journals/corr/abs-1812-02510} and \cite{du2020generalizable}, incremental learning in \cite{9035099}, pre-processing artifacts in \cite{10.1007/978-3-030-31456-9_15}, transferability of the network in \cite{wang2020cnngenerated}, time dimension with attention mechanism in \cite{fernando2019exploiting}. Other works are \cite{Li_2020_CVPR}, \cite{Verdoliva_2019_CVPR_Workshops}, \cite{akhtar2020utility} and \cite{tolosana2020deepfakes}. Most studies have a heavy focus on DNNs/GAN generated artifacts and do not consider other types of manipulations. Also, none of the aforementioned studies take into account the rationality of the attacker nor the case in which the attacker has multiple choices of attack species. A summary of the most representative works in anti counter forensics, presentation attack detection and deepfakes detection is presented in Table \ref{tabel1:summary}.

\begin{sidewaystable*}[htbp]
\centering
\caption[Representative works on anti counter forensics, presentation attack detection and DeepFakes detection]{\label{tabel1:summary}\change{}{Representative works on anti counter forensics, presentation attack detection and DeepFakes detection. AUC = Area Under the Curve; ER = Error Rate; EER = Equal Error Rate; HTER = Half Total Error Rate; Acc = Accuracy.}}
\begin{tabular}{|l|l|c|c|c|}
\hline {\textbf{References}} &  {\textbf{Approach}} & {\textbf{Database}} & {\textbf{Performance}} &  {\textbf{Year}} \\ \hline \hline
\multicolumn{5}{|c|}{\textbf{Anti Counter Forensics}}
\\ \hline 
De Rosa \textit{et al.} \cite{7004806} & Second-order statistics derived from co-occurrence matrices & UCID.v2  & AUC = 90\%  & 2015 \\ \hline
 Barni \textit{et al.} \cite{8081213}& Higher-order features both in spatial and frequency & RAISE & AUC = 98\% & 2017 \\ 
 & domain + SVM & & & \\ \hline
 Chen \textit{et al.} \cite{8653427}& Randomisation of the feature space + SVM & RAISE-2k & ER = 90\% & 2019 \\ \hline
 \multicolumn{5}{|c|}{\textbf{Presentation Attack Detection}}
\\ \hline 
 Boulkenafet \textit{et al.} \cite{7748511} &Speeded-Up Robust Features and Fisher Vector Encoding & Replay-Attack, & EER = 0.1\%, & 2017 \\
 & + Softmax classifier with a cross-entropy loss function &CASIA, MSU   & 2.8\%, 2.2\% &  \\ \hline
 Wang \textit{et al.} \cite{WANG2017332}& CNNs-based texture features + depth information  &   In-House & HTER = 1.2\%& 2017  \\ 
 & from Kinect + SVM & & & \\ \hline
 Liu \textit{et al.} \cite{Liu_2019_CVPR}& Zero-shot Deep Tree Network partitioning spoof samples & CASIA, &AUC = 90\%,  & 2019 \\ 
 & into semantic sub-groups in an unsupervised fashion &  Replay-Attack, MSU & 99.9\%, 81.6\% &  \\ \hline
 Li \textit{et al.} \cite{Li2020compactnet}& Deep-learning system (i.e., CompactNet) for learning  &Replay-Attack, & HTER = 0.7\%, & 2020\\ 
& a compact space tailored for face PAD &OULU-NPU, & 6.0\% &\\ 
& &HKBU-MARs V1&  14.8\% &\\ \hline
 \multicolumn{5}{|c|}{\textbf{DeepFakes Detection}}
\\ \hline 
Li \textit{et al.} \cite{8630787} & Eye-blinking analysis using Long-term Recurrent & CEW, & AUC = 99\% & 2018 \\ 
&  Convolutional Networks & Generated DeepFakes &  &  \\ \hline
Nguyen \textit{et al.} \cite{8682602} & CNNs-based capsule network & FaceForensics & Acc = 83.33\% & 2018 \\ \hline
Du \textit{et al.} \cite{du2020generalizable}& Locality-Aware AutoEncoder & FaceSwap & Acc = 68.06\% & 2020 \\ \hline
 \end{tabular}
\end{sidewaystable*}

\section{Theory}
\label{theory}

In this section, I introduce the definition of a rational attacker and formulate such an attacker's pay-off equation and decision-making process. 
Furthermore, I discuss the detection strategy facing such an attacker and define the requirements for a PAD system accordingly. 
Lastly, I justify the use of one-class detection techniques based on generative models for unknown attack detection.

\subsection{Rational Attacker}
In most existing literature the selection process of the attackers for which attack species to use is neglected and assumed to be that of random selection, resulting in the proposed detectors having fundamental weaknesses. 
A rational attacker is defined as an attacker who, knowing the pay-offs to his possible choices, selects the one with the highest pay-off. 
From a game-theoretic perspective, the interactions between an attacker $x$ and the defender can be modeled by a sequential asymmetric game in which the defender chooses a detector after which the attacker administers their attack of choice. 
An attacker would have to choose among a set of attack species $A_x$ which represents all his options.
The pay-off $u_i$ for the attacker for an attack $a_i \in A_x$ can be formulated as:
\begin{align}\label{attack_payoff}
\begin{split}
u_i & = r (1-p_i) - c_f p_i - c_i \\
    & = r - p_i (r + c_f) - c_i \\
    & \cong - p_i (r + c_f) - c_i
\end{split}
\end{align}
where $r>0$ is the reward for a successful attack, $p_i$ is the probability of detection (detection rate) for the attack species $a_i$, $c_f>0$ is the cost of failure for the attacker, and $c_i>0$ is the cost of the attack.
To account for the budget of the attacker, I assume the budget allows all attack species that are in $A_x$, and any attack that requires a higher budget is excluded from $A_x$.

The attacker can, with the help of trial and error as well as consultation from the experience of other attackers, have an accurate estimate of $p_i$ for $a_i \in A_x$.
The attacker's goal is to choose an attack species that maximizes the pay-off function if the highest pay-off is higher than the pay-off of not attacking the system.
As $r+c_f$ is constant for every individual attacker, the optimization corresponds to the selection of an attack species with the lowest weighted sum depending on $p_i$ and $c_i$.
In practice, it is fruitful for the defender to take $c_i$ into account, and low-cost attack species are expected to occur more frequently than the high-cost ones.
However, because measuring $c_i$ for individual attack species falls outside the scope of this study, I assume the worst-case scenario in which the cost of all possible attack species are assumed zero, enabling all attackers to use more effective attacks regardless of the cost of the attack, as long as their budget allows the attack to be included in $A_x$.
Consequently, the pay-off formula boils down to $u_i \cong -p_i$, and the choice of the attacker would be the attack with the lowest $p_i$, referred to as the \textit{most powerful attack} (MPA).
The values for $p_i$s depends solely on the choice of the detector by the defender.

\subsection{Multiple Attackers}
\label{multiattacker}
A detection system faces not only one attacker but different attackers with different sets of $A_x$.
Gathering statistics about the availability of attack species to the attackers would provide further knowledge about the probability of observing a specific MPA during the detection scenario.
However, as such statistics are often not available for individual attackers, a conservative approach would be to construct a union set of all possible attack species for groups of attackers $A_{X_k}$ and assume all attack species in $A_{X_k}$ are available to all attackers from category $k$.
By doing so, the PAD scenario is further simplified as the distinction between individual attackers collapses and all attackers in each category become identical. 

For example, using the budget as a categorizing factor, the attackers can be categorized to low-budget and high-budget and the attack set for low-budget attackers $A_{X_l}$ and high-budget attackers $A_{X_h}$ can be constructed.
Next, using the probability of an attacker belonging to each category $p({X_k})$ and the performance of the detector $D$ on the MPA from that category $perf({A_{X_k}}|D)$, the expected overall performance of the system can be estimated as $\sum_{k} p({X_k})\times perf({A_{X_k}}|D)$.
Other examples of categorizing factors are expertise, time-budget, and access to unknown attacks or anti-forensic attacks.
As the categorization of the attackers and calculation of the probability of attackers belonging to each category falls outside the scope of this study, I assume a single category $A_X$ for all attackers.
From here on, I use the term \textit{attacker} to refer to the hypothetical attacker that can administer all attacks in $A_X$. 

\subsection{Detection Strategy}
\label{strategy}
For deciding the best detection strategy, the accurate estimate of detection rate for individual attack species by the attacker can be interpreted as equivalent to having full knowledge over the detection performance over all $a_i \in A_X$. 
Due to the sequential nature of the game, the defender needs to choose $p_i$s for individual attack species before the attacker decides which attack to choose.
Subsequently, the rational attacker will choose the MPA which has the lowest detection rate depending on the defender's choice of detector.

\change{Let us assume the set $A$ which denotes all possible attack species. In $A$, two attack species are considered different if they have different manufacturing/generation process, including generation parameters such as manufacturer expertise, quality, and obfuscation. 
From the perspective of an attack detection system, an attack species can be categorized into one of three subsets: (1) Known attack species ($A_k$) to which the detector is exposed in the training process and its performance optimized, (2) Unknown attack species ($A_u$) to which the detector is not exposed to and its performance is unknown, and (3) Anti-forensic attack species ($A_a$) signifying the attack species that are designed with knowledge over the weaknesses of the detector in mind and render the detector useless. These three subsets cover the whole set $A$. It is important to mention that these subsets can be expanded as new attacks are invented (become possible) and added to $A$.}{
Let us assume the set $A$ denotes all possible attack species. In $A$, two attack species are considered different if they have different manufacturing/generation process, including generation parameters such as manufacturer expertise, quality, and obfuscation. 
From the perspective of an attack detection system, an attack species can be categorized into one of three subsets: (1) Known attack species ($A_k$) to which detector is exposed in training process and its performance optimized, (2) Unknown attack species ($A_u$) to which detector is not exposed to and its performance is unknown, and (3) Anti-forensic attack species ($A_a$) signifying the attack species that are designed with knowledge over the weaknesses of the detector in mind and render the detector useless. These three subsets cover the whole set $A$. It is important to mention that these subsets can be expanded as new attacks are invented (become possible) and added to $A$.}

To the extent of the knowledge available to the defender, $A_k$ constitutes the set of all possible attack species, all while the attacker may be able to administer attacks falling outside $A_k$. The defender can know the detection rate for attack species in $A_k$ and optimize them accordingly, however, he cannot know the detection rate for attack species in $A_u$. The best the defender can do in this case is to make an educated guess of what the minimum detection rate can be for attack species in $A_u$. To achieve this, every individual attack species in $A_k$ can be left out as an imaginary unknown attack species during training, and the minimum detection rate across all leave-one-out (LOO) trials can be used as a rough estimate of the detection rate across MPA in $A_u$.

The pay-off for the defender can be formulated as \begin{align}\label{detect_payoff}
\begin{split}
v_i = -c_d - c_m(1-p_i),
\end{split}
\end{align}
where $c_d$ is a constant cost of detection, $c_m$ is the constant cost of missed detection, and $p_i$ is the probability of detection of attack $a_i$ which matches the definition of $p_i$ for the attacker. Knowing that the attacker will choose MPA, i.e. the attack species with the lowest $p_i$, the defender's best strategy would be to maximize the minimum $p_i$ across both $A_k$ and $A_u$ to maximize $v_i$. There is a further objective of reducing the detection cost such that $c_d$ is not prohibitively large, i.e. $c_d << c_m(1-p_i)$. The defender needs to choose to maximize $p_i$ either for $a_i \in A_k$ or $a_i \in A_u$, while limiting $c_d$ according to the application dependant $c_m$. As mentioned in Section \ref{multiattacker}, it is also possible to categorize the attackers to the ones with access to attack species from $A_u$ and the ones without, and define an objective function that takes into account the minimum detection rate over both $A_k$ and $A_u$. Yet, as the defender does not possess any knowledge over $A_u$, it logically follows that he does not have any knowledge about the probability of the attackers being able to use attacks that belong to $A_u$ either, and would need to resort to an educated guess of the probability instead. In this study, I try to maximize the detection rate for MPA from $A_k$ and $A_u$ independently, corresponding to the cases where $A_X \subset A_k$ and $\exists a_i \in A_X, a_i \in A_u$ respectively, and propose a fusion scheme that can be used to combine the resulting detectors without a significant loss of performance in either case.

\subsection{Requirements}
\label{requirements}
Following the aforementioned explanations, it is evident that the common approach towards improving the average detection performance across known attacks is not viable when the detectors are deployed and face rational attackers. Consequently, a more sophisticated approach is needed to be taken based on these analyses where the performance of a system is optimized considering the MPAs, unknown attacks, and adversarial attackers. To this end, the following set of requirements can be defined as guidance for the development of a robust detection system:

\begin{itemize}
    \item It should have an optimal minimum detection rate across known attack species.
    \item It should have an acceptable minimum expected detection rate across unknown attack species.
    \item It should be able to learn to detect an unknown attack species optimally once it becomes known by a few examples.
    \item The cost of detection should not outweigh the cost of miss-detection.
    \item It should be robust against adversarial attacks.
\end{itemize}

The first two requirements can be directly justified according to the formulation of the problem provided in Sections \ref{strategy}. The third requirement follows directly from the first two for the case when an unknown attack species becomes known. In this case, the newly known attack species qualifies for a known attack species and should follow the first requirement, even though there might exist only a limited number of available examples from it. Consequently, the detector should be able to learn to increase the detection rate of the previously unknown attack species to match that of known ones.

\change{There are certain solutions in the literature that attempt to address the last requirement }{}\change{, however, to the best of the author's knowledge, there exists no method to prove the robustness mathematically, and empirical proofs would be limited to the specific anti-CF attacks that are considered. Consequently, for a detector to achieve robustness against adversarial attacks, it needs to survive the test of time. As such, fulfilling this requirement falls outside the scope of this study.}{
There are certain solutions in the literature that attempt to address the last requirement }\cite{8653427}\change{}{, however, to the best of our knowledge, there exists no method to prove the robustness mathematically, and empirical proofs would be limited to the specific anti-CF attacks that are considered. Consequently, for a detector to achieve robustness against adversarial attacks, it needs to survive the test of time. As such, fulfilling this requirement falls outside the scope of this study.}

\subsection{Generation-based Feature Sets}
\label{complete_fs}
\change{It is common practice to rely on discriminative models for detection of attacks. However, the objective of a discriminative model requires it to focus on the discriminative features between bona fide (BF) and known attack species. Consequently, these models do not learn discriminative features that are not directly useful for the detection of the presented known attacks. As such, these models often fail to infer information on unknown attacks where the discriminative feature set is different from the learned ones. In contrast, the objective of a generative model trained on BF data requires it to model all variability in the BF data to the capacity of the model, and because of this, does not over-represent some features while under-representing the others. }{It is common practice to rely on discriminative models for the detection of attacks. However, the objective of a discriminative model requires it to focus on the discriminative features between bona fide (BF) and known attack species. Consequently, these models do not learn discriminative features that are not directly useful for the detection of the presented known attacks. As such, these models often fail to infer information on unknown attacks where the discriminative feature set is different from the learned ones. In contrast, the objective of a generative model trained on BF data requires it to model all variability in the BF data to the capacity of the model, and because of this, does not over-represent some features while under-representing the others. }
\change{Using feature sets extracted by a generative model, a detector would be intrinsically robust to unknown attack species as it has access to the whole feature set, only limited by the capacity of the generator in learning the feature set corresponding to BF data.}{Using feature sets extracted by a generative model, a detector is expected to be more robust to unknown attack species as it has access to more informative feature sets \cite{perera2020generative}, only limited by the capacity of the generator in learning the feature set corresponding to BF data \cite{neal2018open}. Namely, GANs have shown to be more effective for open-set recognition \cite{ditria2020opengan}.} \change{Hence, generative models can be used for anomaly extraction more effectively in the unknown attack detection scenarios.}{Hence, generative models can be used for anomaly extraction more effectively in unknown attack detection scenarios.} Even though the features extracted using the generative model are not optimized for detection and might not outperform the discriminative features used by a discriminative model on known attacks, it can be demonstrated that they would generalize better on unknown attack species as they have no bias regarding what the attack should look like \cite{ditria2020opengan}\cite{neal2018open}\cite{perera2020generative}.

\subsection{Minimax Objective Function}
Considering the known attack detection scenario, another limitation of most existing discriminative detectors is the reliance on the average loss for optimizing the parameters. However, as argued in Section \ref{strategy}, the performance of a detector against a rational attacker is not determined by the average detection rate, but the detection rate on the MPA. Accordingly, optimizing the average detection rate does not necessarily translate to the optimization of the detection rate against the MPA all while posing challenges for the detection of the under-represented attack species. In response to this limitation, objective functions that rely on minimizing the maximum loss (or maximizing the minimum gain) are proposed as a reliable alternative, for which the GAN loss \cite{goodfellow2014generative} is a famous example.

\section{Proposed Method}
\label{method}
According to the requirements defined in Section \ref{requirements}, two separate detection methods are proposed for both scenarios of known and unknown attack detection. Furthermore, a fusion mechanism is introduced to combine the decision of the two detectors for a unified solution with few-shot learning capabilities. Both proposed methods rely on pixel-level generator-based anomaly features and its compact representation extracted to achieve better performance across unknown attack species. For the purpose of known attack detection, a new loss function is introduced which follows the defined objective of maximizing the minimum detection rate. For the purpose of unknown attack detection, I construct a generator-based one-class detector that relies on attack-unspecific anomaly-sensitive information extracted from the detection pipeline.

\subsection{Pixel-Level Probability Distribution Modelling}
\label{pixelllk}
A distribution model for BF images can provide an ideal model for presentation attack detection, as it would be a generative model that contains the complete feature-set and can also provide a single detection score in the form of the likelihood of an observation to the BF distribution. However, due to the complexity of the distribution of BF images, the large amounts of data needed to train such distribution properly, and finally the curse of dimensionality, it is deemed impractical. However, by breaking down the problem into modeling segments of an image rather than the whole image, there exist practical solutions. 

\begin{figure}[htbp]
\centering\includegraphics[width=\columnwidth]{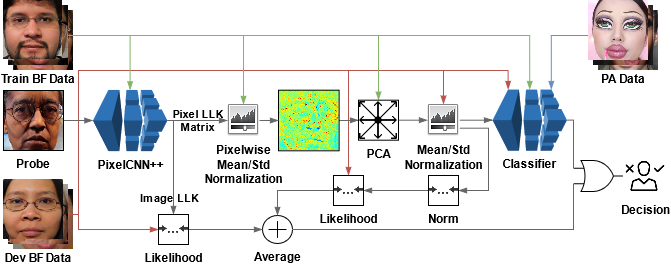}
\caption[The pipeline of the designed detection mechanism]{The pipeline of the designed detection mechanism for both the discriminative classifier and the generator-based one-class detector. The red, green, and blue arrows signify the use of the data in the training of the models pointed to. The gray arrows correspond to the flow of the probe data.}
\label{fig_pipeline}
\end{figure}

PixelRNN \cite{DBLP:journals/corr/OordKK16} is a generative model that models the pixel intensity value probability distribution conditioned on previous pixel values in raster order. This approach can be used to calculate log-likelihood values for observing individual pixels in an image, and once these values are aggregated, they can be used to estimate the log-likelihood of observing the input image as a whole. The pixel-level log-likelihood values can further be used for the localization of low-likelihood pixels (anomalies) in the input. In the proposed approach, the aggregated log-likelihood value is used as the first anomaly measurement for the one-class classifier, and a dimensionality reduction scheme is proposed for simplification of the description of the localization information for extracting the second anomaly measurement which are also used for training the proposed discriminative detector for the known attack detection (Fig. \ref{fig_pipeline}).

\subsection{Dimensionality Reduction}
\label{pca}
The pixel-level log-likelihood values provide valuable information about the severity of the anomalies at each location in the image. However, dealing with features the same size as the input video proves challenging, especially when the amount of training data is limited. To tackle this problem, the following dimensionality reduction scheme is proposed: As the location of anomalies in expected to remain roughly constant in a video, one can average the pixel-level log-likelihood values across the cropped face frames across the whole input video. This step will serve two purposes, firstly it collapses the data in the time dimension, and secondly, it reduces the noise in the frame-level representations. Next, I use a principal component analysis (PCA) transformation learned on BF data to reduce the dimensionality further (Fig. \ref{fig_pipeline}). 

PCA transformation extracts the directions where the variability of the BF data is most explained. It can also be used to extract the directions in which the input data shows little variability. The components for which the BF data shows little variability fits well with the definition of anomaly features, and they are a good representation of the similarities between the BF samples. Additionally, the unexplained variability of input after transformation to the PCA space can provide further anomaly clues. This unexplained variability can be measured as the distance between the input and its projection on the PCA hyper-plane. Thus, I augment the PCA transformed features with the measurement of unexplained variability. The resulting compact representation manages to conserve the discriminative information in the input video effectively while reducing the dimensionality further by a factor of $\approx 1000$.

The amount of shift across the PCA dimensions where BF samples show little variability, along with the unexplained variance measurement can directly be used for one-class detection. To reduce it to a single score, the energy of the input across these dimensions can be measured by calculating the norm of the signal across them. However, as the unexplained variability is on a different scale compared to the PCA transformation values, a normalization step is required. Normalization can be done by making the distribution of the BF samples across these dimensions zero-mean unit-variance.

\subsection{Categorical Margin Maximization Loss}
As the performance of a system in deployment is measured according to its performance for the MPA, a new minimax loss function needs to be introduced that optimizes the detector towards achieving the highest MPA detection rate possible. In this approach, motivated by the success of the triplet loss \cite{schroff2015facenet} , I introduce categorical margin maximization loss (C-marmax) that weighs attacks exponentially according to the difficulty of classification, and thus focuses on reducing the loss from the most difficult samples (MPAs) at each batch during the training. Using C-marmax, the network transforms the aforementioned compact representations to embeddings on a unit hyper-sphere where the distance between the BF data and attacks are maximized while the distance between attacks from the same species, as well as BF samples to each other, is minimized. In this loss, the distances between attacks from one species to other species are ignored as we don't have any information about the similarity or dissimilarity between distribution across any two attack species. Hence the detector is \textit{categorical} as it only considers distances between observations from different categories (i.e. BF vs attacks) for calculating the loss value. Finally, to exaggerate the loss from samples belonging to the MPA and suppress the loss from other attack species, the loss attributed to the anchors is exaggerated according to the distances such that the network pays more attention to marginal anchors to fulfill the objective of maximizing the minimum detection rate.

In attack detection scenarios, there are a few classes, and it is possible to rely on the distance to the center of distribution in a batch rather than the distance between individual samples. To this end, in each batch, I compute the location of the center of distribution for each attack species as well as BF data on the unit hyper-sphere, and according to the label of the inputs, I use these centers to measure the distance of the anchor to the positive distribution $d_p$ and the negative distribution $d_n$. To achieve the maximum margin possible between the distribution of BF samples and PA samples in the embedding space, a fixed margin is not defined. Instead, the ratio $\frac{d_p}{d_n}$ is used for the maximum $d_p$ and minimum $d_n$ in a batch from each class, requiring the numerator to be minimized to zero, while the denominator is maximized to the maximum possible value of $2$ on the unit hyper-sphere. To avoid the loss value to become infinity when $d_n$ is zero, the ratio is modified to $\frac{d_p}{d_p+d_n}$ which is equivalent to $\frac{d_p}{d_n}$ when $d_p << d_n$. Furthermore, to exaggerate the loss for marginal observations (where $d_p$ is high) in comparison to non-marginal observations (where $d_p$ is low), exponentiation is used, and the resulting formula becomes $(\frac{d_p}{d_p+d_n})^g$. 

As the defined loss does not maximize the distance between centers of distributions directly, to assure that the center of distributions are far from each other, the minimum distance between two centers are floored at $\sqrt{2}$ corresponding to $90$ degrees on the unit hyper-sphere, with a second loss term. The final loss function is summarized as follows:
\begin{align}\label{loss}
\begin{split}
loss_m &= (\frac{max\{d(a, C_p)\}}{(max\{d(a, C_p)\}+min\{d(a, C_n))\}})^g\\
loss_c &= max\{min\{\sqrt{2} -d(C_p, C_n)\}, 0\}\\
loss &= loss_m + 0.1\times loss_c
\end{split}
\end{align}
where, $d$ stands for euclidean distance, $a$ signifies the anchor, $C_p$ is the center of the positive class, $C_n$ is the center of the negative class, $g$ is the exaggeration factor, $loss_m$ is the margin loss, and $loss_c$ is the center loss. During decision making, the euclidean distance to the center of BF distribution can be used for scoring. This distance can further be converted to an attack detection probability value by division by $2$.

In comparison to the triplet-loss, the proposed modifications result in a tunable exaggeration of the loss in misclassified samples and suppression of the loss in the correctly classified ones and relax the need for a fixed margin constant. Having no constant margin, the network can continue training even after a specific margin is achieved between the classes until the maximum margin on the hyper-sphere is reached. Furthermore, by using the center of the distribution instead of the distance between individual anchors, the loss becomes less stochastic, allowing faster convergence. To the same effect, the categorical nature of the proposed loss relaxes the untrue assumption that all attacks come from the same distribution regardless of their corresponding attack species.

\subsection{Unknown Attack Detection}
\label{blind}
As argued in Section \ref{complete_fs}, a discriminative model may overfit to certain discriminative features that correspond to the bias in known attack species used in training. This also holds true for the presented C-marmax loss, as even though it tries to achieve a balanced attack detection performance across known attack species, it may exclude discriminative features that may be important for the detection of unknown attack species. As such, to detect unknown attacks, a one-class detector is proposed which does not have a bias towards any specific attack species, or in other words, for it all attacks are unknown. As explained in Section \ref{pixelllk}, the log-likelihood value of observing an image serves as a good general-purpose anomaly detection measure. However, this metric does not include the other important discriminative feature available in the pixel-level log-likelihood data, namely the location information. As explained in Section \ref{pca}, the location relevant anomalies can be represented by the components in a PCA transform trained on BF where the BF data show the least variability. Furthermore, this representation can be augmented by the unexplained variance in the form of the distance of an observation to the PCA hyper-plane. Finally, the energy of the signal across the resulting representation after normalization can be used as an anomaly score. Following these steps, a second location-sensitive anomaly measure is derived. Assuming a Gaussian distribution for BF scores for both anomaly measures, using the BF score distribution, one can calculate the likelihood of an observation belonging to this distribution as the final probability score. For the final score of the one-class detection scheme, I simply average the two resulting likelihood scores from the log-likelihood measure and the PCA-based measure (Fig. \ref{fig_pipeline}).

To fuse the probability scores from the discriminative detector and the one-class detector when they are employed together, I use the following logic: If the discriminative detector decides that a sample is an attack, it most certainly is one. However, if the discriminative detector decides that the sample is a BF, the defender cannot be sure that the sample is a BF as it might come from an unknown attack. So the one-class detector is to be consulted for a decision. This two step decision logic can be interpreted as using an $OR$ gate on the decision of the discriminative and the one-class detector decisions. However, as both systems provide a probability scores rather than a decision, considering that $A \lor B = A+B-AB = \overline{\overline{A}\times\overline{B}}$, the following fusion formula is proposed that mirrors the logic level decision making:
\begin{align}\label{fusion}
\begin{split}
p_{PA}(x|D, O) &= 1-p_{BF}(x|D)\times p_{BF}(x|O)\\
               &= 1-(1-p_{PA}(x|D))\times(1-p_{PA}(x|O))
\end{split}
\end{align}
where $p_{PA}$ corresponds to the probability of belonging to the attack category, $p_{BF}$ corresponds to the probability of belonging to the BF category, and $O$ and $D$ correspond to one-class and discriminative detector models.

\section{Experiment Setup}
\label{setup}
For measuring the effectiveness of the proposed method, its application on both tasks of presentation attack detection and Deepfake detection are considered. In this section, a description of the datasets used is provided, followed by the parameters used in training. Lastly, the measures used for evaluation of the method are described.

\subsection{Datasets}
To show the performance of the proposed method for presentation attack detection, the SiW-M dataset\footnote{\href{https://web.archive.org/web/20200703043312/http://cvlab.cse.msu.edu/siw-m-spoof-in-the-wild-with-multiple-attacks-database.html}{http://cvlab.cse.msu.edu/siw-m-spoof-in-the-wild-with-multiple-attacks-database.html}} \cite{Liu_2019_CVPR} is selected due to its large collection of presentation attack species. Similarly, the FaceForensics++ dataset\footnote{\href{https://github.com/ondyari/FaceForensics}{https://github.com/ondyari/FaceForensics}} \cite{Rossler_2019_ICCV} is chosen for the task of Deepfake detection as it contains the widest choice of species between the available datasets.

\subsubsection{SiW-M}
This dataset consists of $660$ BF videos from $493$ subjects from diverse ethnicity and age. Furthermore, it includes $966$ PA videos from $13$ different PAS collected under various environmental conditions, extreme face pose angles and lighting conditions. The videos are around six seconds in length.
This dataset is specifically designed for the evaluation of generalization performance across unknown PAS. 
The attack species in this dataset are categorized into replay, print, mask, makeup, and partial attacks. 
The PAS available in this dataset are form a diverse set of attacks including print and display attacks as well as transparent masks and impersonation makeup. This dataset also includes PAS corresponding to partial attacks.

For training the models, $530$ randomly chosen BF videos are used, while $65$ randomly chosen BF videos were kept for development purposes, leaving $65$ videos for testing. For training the classifier in the unknown case, a LOO setup is used and for each attack species, all the videos from other attack species are used for training, along with the training and development BF data. 
For few-shot learning, an additional randomly chosen one or five videos from the targeted attack species are included in the training, while in the known case $50\%$ of the videos are included.

\subsubsection{FaceForensics++}
FaceForensics++ dataset contains four PAS corresponding to Deepfakes\footnote{\href{https://github.com/deepfakes/faceswap}{https://github.com/deepfakes/faceswap}}, Face2Face \cite{Thies_2016_CVPR}, Faceswap\footnote{\href{https://github.com/MarekKowalski/FaceSwap/}{https://github.com/MarekKowalski/FaceSwap/}}, and Neural Textures \cite{10.1145/3306346.3323035}. The dataset contains $1,000$ BF videos and $1,000$ videos from each PAS, each split into three sets, reserving $72\%$ for training, $14\%$ for validation and allocating $14\%$ for evaluation. The videos are collected from YouTube and after manipulation, recompressed in three video qualities for evaluation of performance under various compression levels. For the purpose of analyzing performance over unknown attacks, only the non-compressed version of the data is used. Similar to the SiW-M dataset, both known and LOO unknown attack detection experiments are considered.

\subsection{Parameters}
The proposed method has a number of parameters corresponding to face detection, the pixel-level log-likelihood extraction model, the PCA model, and finally the classifier. In this study, the videos are considered as a set of frame images. The face region is extracted in each frame after face detection using the Dlib toolkit \cite{king2009dlib}, and the cropped faces are resized to $128\times128$.

The overall pipeline of the proposed detection mechanism is visualized in Fig. \ref{fig_pipeline} along with information about where the training data, development data, and known attack data is used. The input image is first processed by the PixelCNN++ model trained using the training data, resulting in an aggregated observation log-likelihood and pixel-wise log-likelihood matrices. The aggregated observation log-likelihood is compared to the distribution of BF values learned from development data to acquire the first generator-based anomaly measure. The pixel-wise log-likelihood matrices are further normalized to zero-mean unit-variance using the distribution of pixel values in the training data before applying the PCA transform. The PCA transform is learned using the training data, and the PCA transformed representation is augmented with the unexplained variance measure and normalized to zero-mean unit-variance across all dimensions using the development data. Then after sorting the components based on the explained variance of training data in descending order, the last components are used for calculating the norm. This value is then compared to the distribution of BF scores learned on development data for calculating the second generator-based anomaly measure. The first and second probability scores are combined by averaging, resulting in a single one-class classification score. The augmented and normalized PCA representations are then passed to the discriminative classifier trained on BF data from training and development set along with attack data from known attacks.

\subsubsection{PixelCNN++}
For pixel-level log-likelihood matrix extraction, a PixelCNN++\footnote{\href{https://github.com/openai/pixel-cnn}{https://github.com/openai/pixel-cnn}} \cite{DBLP:journals/corr/SalimansKCK17} model is trained on the resized cropped face images extracted from the BF training data. The model consists of three hierarchies with five ResNet layers in each, with $160$ filters with a receptive field of $3\times3$ in each layer, resulting in $95$ million parameters. Concatenated ELU \cite{10.5555/3045390.3045624} is used for activation and pixel intensity values are modeled using $10$ logistic distributions. For regularization, dropout with a probability of $50\%$ is used. The model is trained with a batch size of one and the ADAM \cite{DBLP:journals/corr/KingmaB14} optimizer with a learning rate of $10^{-5}$ is used for $500$ epochs on a single randomly chosen frame per training video in each epoch.

The log-likelihood matrix is then generated by concatenating the pixel log-likelihood values for each of the $10$ logistic distributions for each color channel, resulting in a matrix of size $128\times128\times30$. For calculating the log-likelihood of observing the video, the likelihood of observing each individual frame is calculated using the weighted sum for the individual logistic distributions across the whole cropped face image. These values are then averaged across time to measure the average log-likelihood of the observed input video to be used for one-class detection. For extracting location-sensitive features, after averaging the pixel-level log-likelihood matrix values across the whole input video, at each pixel location, the distribution of log-likelihoods are normalized such that the BF training data has a distribution of zero-mean unit-variance, resulting in a matrix of size $128\times128\times30$ per video. 

\subsubsection{Principal Component Analysis}
In the next step, these matrices are extracted from the BF training data to train a PCA model with sorted components according to the explained variance across these components in descending order. Unexplained variance is measured by calculating the euclidean distance between each input and its projection on the PCA hyper-plane and added to the end of the PCA representation. The PCA representation is normalized to have zero-mean unit-variance for BF data from the validation set. For one-class detection, to measure the energy of the input video across the last $10\%$ of the PCA representation, the norm after normalization is used. Using the distribution of the norm values across the validation data, a single Gaussian model is trained for calculating the likelihood of a given input to the BF distribution. The same approach is taken for the video log-likelihood values collected directly from the output of the PixelCNN++ model. These two likelihood values are averaged to calculate the final score of the generator-based one-class detector.

\subsubsection{Classifier}
\begin{figure}[htbp]
\centering\includegraphics[width=\columnwidth]{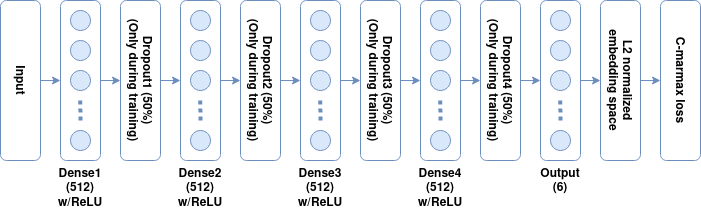}
\caption[The architecture of the classifier network]{The architecture of the classifier network.}
\label{fig_network}
\end{figure}

The PCA representation is also used for the training of the discriminative classifier using the aforementioned loss function. A DNN model with four hidden layers, each with $512$ ReLU activated units is trained for mapping its input to the L2 normalized embedding space of six dimensions (Fig. \ref{fig_network}). Due to the limited amount of training data available for training the classifier, dropout regularization with a rate of $50\%$ is used on the output of each hidden layer, along with L2 regularization with a factor of $10^{-6}$. Oversampling is done by using random segments of the training videos and their vertically flipped copies while testing is done on the whole test videos. The training data is balanced by repetition to have $50\%$ BF samples and $\frac{50\%}{\#PAS}$. The loss function only has one tunable parameter $g$, which was set to two to achieve fast conversion. Training is done with a batch size of $128$ for $100$ epochs with a fixed learning rate of $10^{-3}$ using the ADAM optimizer. Finally, the detection probability score is calculated by measuring the Euclidean distance of the embedding to the average of the validation data embeddings divided by two. The fusion between the probability score calculated by the generator-based one-class detector and the discriminative detector is done using the formula in Section \ref{blind}.

\subsection{Metrics}
To evaluate the performance of the proposed system, the threshold less equal-error-rate (EER) metric is used. EER measures the error rate when the missed detection percentage is equal to the false alarm percentage. 
For evaluation of performance across all attack species, the EER value for the MPA is chosen by measuring the maximum EER across all species following the arguments represented in Section \ref{strategy}. Furthermore, the detection error trade-off (DET) curve is used for showing the missed detection rate for each false alarm value. Missed detection corresponds to the bona fide presentation classification error rate (BPCER) and false alarm corresponds to attack presentation classification error rate (APCER) in ISO/IEC 30107 terminology\footnote{\href{https://www.iso.org/obp/ui/iso}{https://www.iso.org/obp/ui/iso}}.\change{}{In the experimental result tables, we report are ACER@APCER=5\%. The BPCER@APCER = 5\% can be calculated as BPCER = (ACERx2)-5\%}.

\section{Presentation Attack Detection}
\label{pad}

In this section, the adequacy of the proposed generator-based anomaly representations is first explained. Later, the performance of the proposed method based on these representations is evaluated and compared to the existing solutions in both known and unknown attack detection scenarios. Lastly, the few-shot learning capacity of the proposed method is investigated and the computational cost of the pipeline is reported.

\subsection{Representation Adequacy}
\begin{figure*}[htbp]
\centering\includegraphics[width=\textwidth]{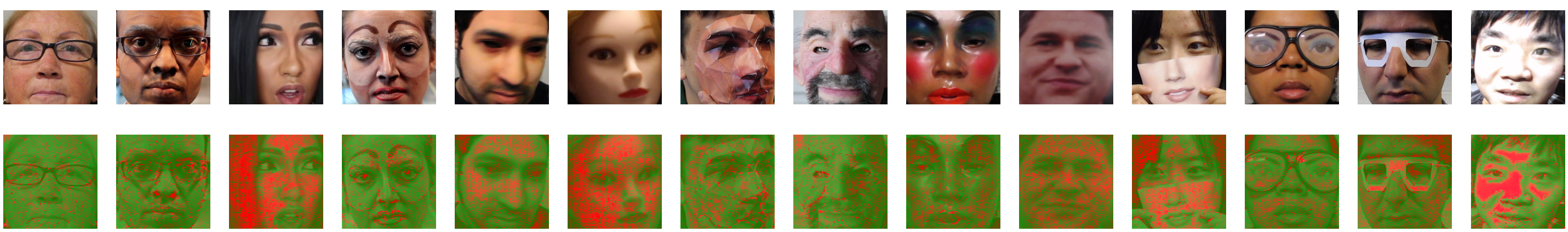}
\caption[Example frames from BF and each PAS from the SiW-M dataset along with their corresponding log-likelihood matrices]{Example frames from BF and each PAS from the SiW-M dataset along with their corresponding log-likelihood matrices below them. Red pixels show the location of anomalies from the perspective of the PixelCNN++ model. From left to right: BF, Cosmetic Makeup, Impersonation Makeup, Obfuscation Makeup, Half Mask, Mannequin, Paper Mask, Silicone Mask, Transparent Mask, Print, Paper Cut, Funny Eye, Paper Glasses, and Replay.}
\label{PAD_examples}
~\\
\centering\includegraphics[width=\textwidth]{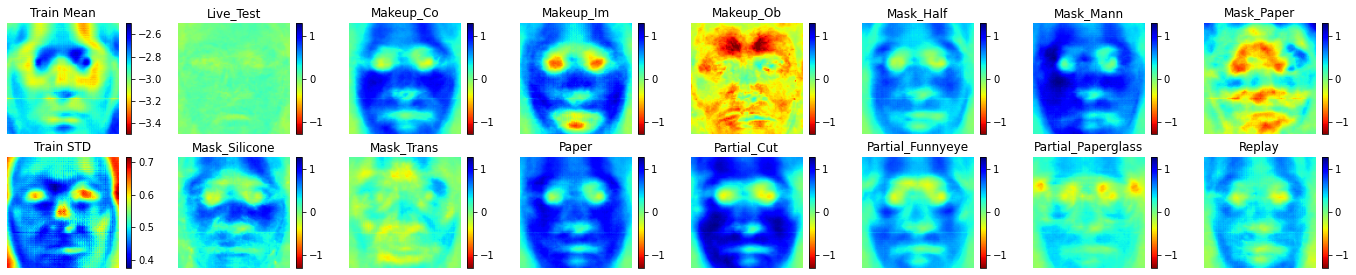}
\caption[Average and standard deviation of the log-likelihood matrices from the SiW-M dataset]{Average and standard deviation of the log-likelihood matrices over training data in the first column, along with the average log-likelihood matrices for test BF data and each individual PAS from the SiW-M dataset in the same order as in Fig. \ref{PAD_examples}.}
\label{average_PAD}
\end{figure*}

Fig. \ref{PAD_examples} shows examples of the log-likelihood matrices extracted by the PixelCNN++ model for sample frames from BF data as well as each attack species. It can be seen that BF data shows few single anomaly pixels corresponding to the natural variations in the BF frame as well as anomalies around the location of the glasses. However, each attack species shows its own pattern of anomalies corresponding to the locations where it is observed. For example for the obfuscation makeup attack, the anomalies correspond to where the eyebrow and beard lines are drawn, for the mannequin attack they correspond to the skin regions, for the paper mask to the fold locations, and for the replay attack to the overexposed regions of the face. These examples show the capacity of the representation to provide explainability at pixel-level.

To further analyze the unique patterns from each attack species, the average log-likelihood matrix for each species is presented in Fig. \ref{average_PAD}. The average and standard deviation of log-likelihood values for training BF data are shown in the first column. From these two images, it can be seen that most of the natural variability in the training data corresponds to the eye and the nasal dorsum as well as the background, while the periocular region of the face contains a lower natural log-likelihood. After normalization of the average log-likelihood matrices for test data using these two matrices, it can be seen that the test BF data matches the training BF data average, while each attack species show a different pattern for low likelihood and high likelihood regions. Attacks with unusually high likelihood over the skin region are cosmetic makeup, impersonation makeup, half mask, mannequin, silicone mask, print, and partial cut attacks. This effect can be interpreted as the over-smoothness of skin texture in these attacks. Attacks with unusually low likelihood over the skin are obfuscation makeup, paper mask and to some extent transparent mask, which can, in turn, be interpreted as severe anomalies in the skin texture. As expected, partial attacks show anomalies in the region of the image where the attack is applied to.

\begin{figure}[htbp]
\centering\includegraphics[width=0.7\columnwidth]{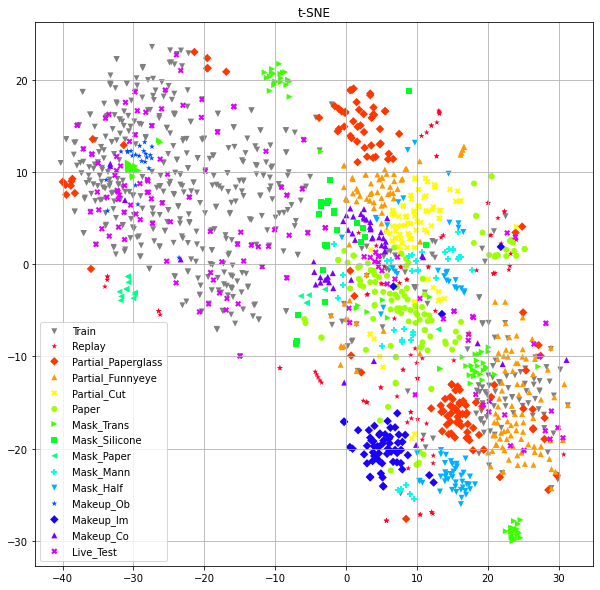}
\caption[The t-SNE graph on the average log-likelihood matrices for all the data available in the SiW-M dataset]{The t-SNE graph on the average log-likelihood matrices for all the data available in the SiW-M dataset. Each point represents a video, and each attack species is visualized with a different shape and color. The training BF data is shown with gray dots while the test BF data is shown with pink pluses. A clear separation is visible between BF data and attack data.}
\label{tsne_pad}
\end{figure}

Fig. \ref{tsne_pad} shows the t-SNE embeddings \cite{10.5555/2968618.2968725} of the normalized average pixel log-likelihood matrices from each video. From this figure, it is evident that the representation manages to cluster attacks from the same species together with few exceptions. Furthermore, it shows a good separability between BF data and presentation attack data, while the training BF data distribution overlaps with the test BF data. These are remarkable characteristics for the features generated by the proposed anomaly extraction which was trained in an unsupervised manner on only BF examples. This separation is however not perfect, as a cluster of BF samples are located inside the attack distribution with high overlap with partial funny eye and partial paper glass attacks. In addition, clusters of presentation attacks exist inside the BF distribution. The majority of these samples are from transparent mask, obfuscation makeup, and partial paper glass attacks. By looking at Fig. \ref{average_PAD} it can be seen that all these attacks have a shared characteristic where the average log-likelihood matrix has lower values on the skin region in contrast to other attacks where the skin region shows higher log-likelihood values.

\subsection{One-class classification}
The performance of both anomaly measures in the proposed one-class classification scheme, along with the combined one-class detection score for each species is presented in Table \ref{table_one_class}. Even though the EER values for the detection of individual attacks, with the exception of impersonation makeup attack, are far from acceptable, these anomaly measures show a balanced performance across all attack species. For all attacks, it can be seen that the fusion of these two anomaly measures successfully reduces the EER close to the smaller value of the two, and subsequently the MPA EER is reduced by $10\%$. The method performs significantly better on impersonation makeup attacks compared to the other attacks while transparent mask and paper glasses attacks are the most challenging for the system.

\begin{sidewaystable*}[htbp]
\caption[Detection performance for each of the anomaly measures and their combination on the SiW-M dataset]{Detection performance for each of the anomaly measures and their combination on the SiW-M dataset.}
\label{table_one_class}
\centering
\begin{tabular}{|l|l|r|r|r|r|r|r|r|r|r|r|r|r|r|r|r|}
\hline
\multirow{2}{*}{Method} & \multirow{2}{*}{Metric [\%]} & \multicolumn{1}{c|}{\multirow{2}{*}{Replay}} & \multicolumn{1}{c|}{\multirow{2}{*}{Print}} & \multicolumn{5}{c|}{Mask}                                                                                                                         & \multicolumn{3}{c|}{Makeup}                                                        & \multicolumn{3}{c|}{Partial}                                                                         & \multicolumn{1}{c|}{\multirow{2}{*}{MPA}} \\ 
                              & & \multicolumn{1}{c|}{}                        & \multicolumn{1}{c|}{}                       & \multicolumn{1}{c|}{Half} & \multicolumn{1}{c|}{Silicone} & \multicolumn{1}{c|}{Trans.} & \multicolumn{1}{c|}{Paper} & \multicolumn{1}{c|}{Mann.} & \multicolumn{1}{c|}{Obf.} & \multicolumn{1}{c|}{Imp.} & \multicolumn{1}{c|}{Cosm.} & \multicolumn{1}{c|}{FunnyEye} & \multicolumn{1}{c|}{P.Glasses} & \multicolumn{1}{c|}{P.Cut} & \multicolumn{1}{c|}{}                         \\ \hline
Agg. Log- & EER                    & 22.08                                        & 22.23                                       & 19.57                     & \textbf{23.99}                & 37.69                       & 29.86                      & 19.09                      & 24.41                     & 16.53                     & 25.88                      & \textbf{24.00}                 & 37.08                              & 22.23                          & 37.69                                         \\ 
\rowcolor{gray!50} \cellcolor{gray!0} Likelihood & ACER & 23.01 & 13.93 & 13.38 & 27.19 & 50.57 & 26.43 & 13.11 & 17.23 & 13.88 & 26.48 & 22.95 & 35.33 & 13.51 & 50.57 \\ \hline
Anomaly  & EER                 & \textbf{13.96}                               & 17.64                                       & 16.67                     & 29.97                         & \textbf{23.86}              & \textbf{23.13}             & 20.34                      & 37.85                     & \textbf{1.52}             & \textbf{22.39}             & 33.54                          & 27.02                              & 18.97                          & 37.85                                         \\ 
\rowcolor{gray!50} \cellcolor{gray!0} Energy & ACER & 17.70 & 27.56 & 14.14 & 31.73 & 33.90 & 23.40 & 15.38 & 31.62 & 4.79  & 36.33 & 31.29 & 30.03 & 15.03 & 36.33 \\ \hline
Combination & EER                        & 15.23                                        & \textbf{12.47}                              & \textbf{14.46}            & 25.84                         & 27.08                       & 23.89                      & \textbf{11.55}             & \textbf{22.99}            & 3.15                      & 23.12                      & 24.18                          & \textbf{26.66}                     & \textbf{15.13}                 & \textbf{27.08}                                \\ 
\rowcolor{gray!50} \cellcolor{gray!0} & ACER & 13.16 & 10.90 & 11.11 & 23.40 & 30.87 & 18.09 & 9.32  & 17.98 & 4.79  & 19.67 & 16.14 & 23.97 & 11.24 & 30.87 \\ \hline
\end{tabular}
\vspace{10pt}
\caption[Performance comparison on the task of known attack detection on the SiW-M dataset]{Performance comparison between proposed detection method and existing methods on the task of known attack detection on the SiW-M dataset.}
\label{known_pad}
\centering
\begin{tabular}{|l|l|r|r|r|r|r|r|r|r|r|r|r|r|r|r|}
\hline
\multirow{2}{*}{Method} & \multirow{2}{*}{Metric {[}\%{]}} & \multicolumn{1}{c|}{\multirow{2}{*}{Replay}} & \multicolumn{1}{c|}{\multirow{2}{*}{Print}} & \multicolumn{5}{c|}{Mask}                                                                                                                         & \multicolumn{3}{c|}{Makeup}                                                        & \multicolumn{3}{c|}{Partial}                                                                         & \multicolumn{1}{c|}{\multirow{2}{*}{MPA}} \\
                              & & \multicolumn{1}{c|}{}                        & \multicolumn{1}{c|}{}                    & \multicolumn{1}{c|}{Half} & \multicolumn{1}{c|}{Silicone} & \multicolumn{1}{c|}{Trans.} & \multicolumn{1}{c|}{Paper} & \multicolumn{1}{c|}{Mann.} & \multicolumn{1}{c|}{Obf.} & \multicolumn{1}{c|}{Imp.} & \multicolumn{1}{c|}{Cosm.} & \multicolumn{1}{c|}{FunnyEye} & \multicolumn{1}{c|}{P.Glasses} & \multicolumn{1}{c|}{P.Cut} & \multicolumn{1}{c|}{}                         \\ \hline
Auxiliary \cite{Liu_2018_CVPR}~~ & EER                     & 4.7                                          & \textbf{0.0}                                & 1.6                       & 10.5                          & 4.6                         & 10.0                       & 6.4                        & 12.7                      & \textbf{0.0}              & 19.6                       & 7.2                            & 7.5                                & \textbf{0.0}                   & 19.6                                          \\ 
\rowcolor{gray!50} \cellcolor{gray!0} & ACER & 5.1  & 5.0  & 5.0  & 10.2 & 5.0  & 9.8  & 6.3 & 19.6 & 5.0 & 26.5 & 5.5  & 5.2  & 5.0  & 26.5 \\ \hline
LLIG \cite{deb2020look} & EER                         & \textbf{3.5}                                 & 3.1                                         & \textbf{0.1}              & 9.9                           & \textbf{1.4}                & \textbf{0.0}               & 4.3                        & \textbf{6.4}              & 2.0                       & 15.4                       & \textbf{0.5}                   & \textbf{1.6}                       & 1.7                            & 15.4                                          \\
\rowcolor{gray!50} \cellcolor{gray!0} & ACER & 3.5  & 3.1  & 1.9  & 5.7  & 2.1  & 1.9  & 4.2 & 7.2  & 2.5 & 22.5 & 1.9  & 2.2  & 1.9  & 22.5 \\ \hline
One-class & EER                         & 15.7                                         & 9.6                                         & 12.4                      & 28.7                          & 27.7                        & 22.5                       & 10.3                       & 18.2                      & 3.9                       & 22.9                       & 22.6                           & 26.2                               & 17.6                           & 28.7                                          \\ 
\rowcolor{gray!50} \cellcolor{gray!0} & ACER & 13.2 & 10.9 & 11.1 & 23.4 & 30.9 & 18.1 & 9.3 & 18.0 & 4.8 & 19.7 & 16.1 & 24.0 & 11.2 & 30.9 \\ \hline
C-marmax & EER                      & 9.7                                          & 5.6                                         & 1.5                       & 6.6                           & 4.5                         & 3.0                        & \textbf{3.8}               & 8.0                       & 3.1                       & \textbf{7.8}               & 5.7                            & 7.7                                & 6.5                            & 9.7                                           \\ 
\rowcolor{gray!50} \cellcolor{gray!0} &  ACER & 10.6 & 5.6  & 4.3  & 6.6  & 8.0  & 8.6  & 6.3 & 13.3 & 4.7 & 9.3  & 6.3  & 9.2  & 6.5  & 13.3 \\ \hline
Fusion & EER                        & 6.1                                          & 7.3                                         & 3.6                       & \textbf{4.5}                  & 4.5                         & 3.8                        & 4.8                        & 8.0                       & 1.5                       & \textbf{7.8}               & 4.3                            & 8.5                                & 3.4                            & \textbf{8.5}                                  \\ 
\rowcolor{gray!50} \cellcolor{gray!0} & ACER & 9.1  & 7.1  & 7.3  & 11.9 & 10.2 & 11.6 & 7.0 & 11.7 & 4.7 & 12.3 & 10.8 & 10.7 & 7.3  & 12.3 \\ \hline
\end{tabular}
\end{sidewaystable*}

\begin{figure}[htbp]
\centering\includegraphics[width=0.6\columnwidth]{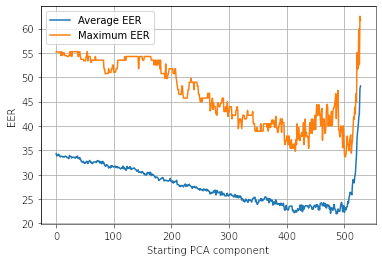}
\caption[Detection performance according to the starting PCA component]{Detection performance according to the starting PCA component before calculation of the energy.}
\label{PCA}
\end{figure}

\begin{figure}[htbp]
\centering\includegraphics[width=0.7\columnwidth]{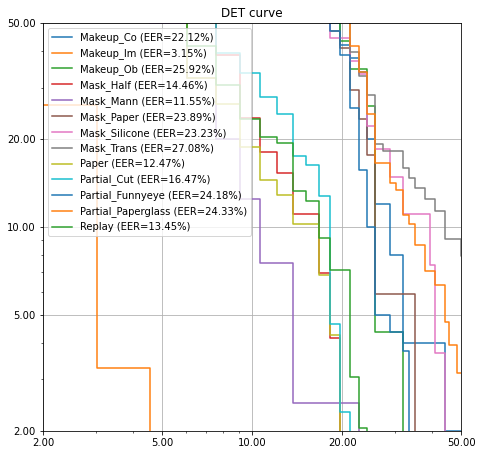}
\caption[Detection error trade-off curve for the one-class detector in PAD on the SiW-M dataset]{Detection error trade-off curve for the one-class detector in PAD on the SiW-M dataset.}
\label{one_class_det}
~\\~
\centering\includegraphics[width=0.7\columnwidth]{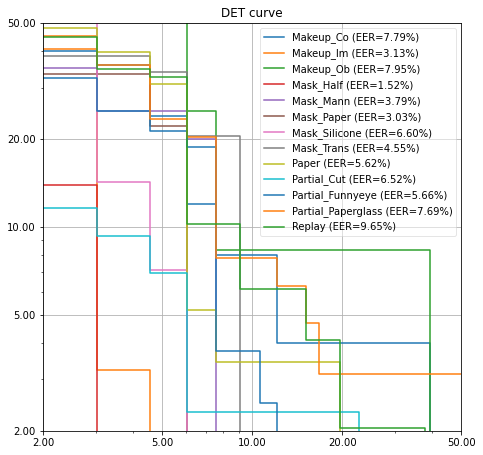}
\caption[Detection error trade-off curve of the discriminative detector for the known attack detection on PAD task on the SiW-M dataset]{Detection error trade-off curve of the discriminative detector for the known attack detection on PAD task on the SiW-M dataset.}
\label{known_det}
\end{figure}

To see the effect of the number of PCA components in the detection rate, Fig. \ref{PCA} shows the average as well as the maximum EER over all species after filtering out the first $n$ components from the PCA representation. It can be seen that, as hypothesized, the last PCA dimensions contain a significant amount of attack-unspecific discriminative information. The correlation between the aggregated log-likelihood measure and the anomaly norm measures is $0.15$ signifying the complementing potential of these measures on each other. The combination scores reflect the complementary nature of these measures and results in a detector with an MPA attack detection EER of $27.1\%$. The DET curve for the resulting one-class detector is shown in Fig. \ref{one_class_det} for all attack species. This plot reveals three clusters of curves corresponding to transparent mask, silicone mask, partial paper glass, paper mask, partial funny eye, obfuscation makeup, and cosmetic makeup attacks with above $20\%$ EERs, partial paper cut, replay, half mask, print, and mannequin attacks with EERs between $10\%$ and $20\%$, and finally impersonation makeup with less than $5\%$ EER. The attacks with higher than $20\%$ EER reflect the overlaps observed in Fig. \ref{tsne_pad}. The attacks with an EER below $20\%$ show similarities in their average log-likelihood images in Fig. \ref{average_PAD} while the other attacks each have their individual dissimilar patterns.

\subsection{Detection Performance}
In the following, the detection performance in terms of MPA EER is presented and analyzed for the detection of known attacks, unknown attacks, and few-shot learning.

\subsubsection{Known attacks}
The performance of the proposed methods in comparison to the existing detection methods which are applied to the SiW-M dataset is reported in Table \ref{known_pad}. It can be seen that even though the proposed method is outperformed on most individual attacks, the focus of the loss function on the MPA resulted in a lower EER on the difficult attack species, namely cosmetic makeup. As a result, the proposed discriminative detector achieves $9.7\%$ EER on the MPA, reducing the MPA EER by $37\%$ compared to the best existing detector. The proposed fusion mechanism further reduces the MPA EER to $8.5\%$. The DET curve for the proposed discriminative detector is shown in Fig. \ref{known_det}. It is worth noting that the clusters visible in Fig. \ref{one_class_det} are merged together and the curves follow a similar course, representing a more balanced detection performance. Furthermore, the curve for impersonation makeup is almost identical to the one-class classification curve, showing that the proposed C-marmax loss successfully avoided optimization of performance on this attack which was the easiest to detect using its input. A similar pattern is observable in Table \ref{known_pad} where print and impersonation makeup attacks achieved the smallest boost in performance after the application of the discriminative classifier. Due to the small number of test samples, the DET curve shows abrupt changes, showing that more data is needed for a more precise measurement of EERs.

\subsubsection{Unknown Attacks}
The results for the proposed method along with the performance of existing detectors in unknown attack conditions are presented in Table \ref{pad_unknown}. It can be seen that, as expected, the one-class detector performs better than all discriminative detection methods in terms of MPA EER, including the proposed method. However, it is worth mentioning that the discriminative detectors gain an advantage over certain PASs where there is a similarity of the discriminative features between the unknown PAS and the known ones used in training. This distinction is visible in cases where a significant difference exists between the one-class classifier performance compared to the discriminative classifier such as in the case of silicone mask and mannequin attacks. A close observation of Fig. \ref{tsne_pad} reveals that samples from these two attacks are not clustered together in the anomaly feature space. The proposed fusion method managed to cap the EER for partial paper glasses and partial funny eye attack where the proposed discriminative detection method fails while not hindering the performance on cases where the discriminative detection method performs well. Considering the existing solutions, it can be seen that there only exists one approach that has a better than chance detection rate for MPA, namely LLIG \cite{deb2020look}. This is concerning as it shows that all other existing methods would be ineffective against a rational attacker, and in the case of \cite{Liu_2018_CVPR}, would actually increase the efficacy of attacks. The proposed method achieves an MPA EER of $27.8\%$ and outperforms all baseline methods.

\begin{sidewaystable*}[htbp]
\caption[Performance comparison on the task of unknown presentation attack detection on the SiW-M dataset]{Performance comparison of the proposed methods and the existing methods in the literature on the task of unknown presentation attack detection on the SiW-M dataset.}
\label{pad_unknown}
\centering
\begin{tabular}{|l|l|r|r|r|r|r|r|r|r|r|r|r|r|r|r|}
\hline
\multirow{2}{*}{Method} & \multirow{2}{*}{Metric {[}\%{]}} & \multicolumn{1}{c|}{\multirow{2}{*}{Replay}} & \multicolumn{1}{c|}{\multirow{2}{*}{Print}} & \multicolumn{5}{c|}{Mask}                                                                                                                         & \multicolumn{3}{c|}{Makeup}                                                        & \multicolumn{3}{c|}{Partial}                                                                         & \multicolumn{1}{c|}{\multirow{2}{*}{MPA}} \\
                              & & \multicolumn{1}{c|}{}                        & \multicolumn{1}{c|}{}                       & \multicolumn{1}{c|}{Half} & \multicolumn{1}{c|}{Silicone} & \multicolumn{1}{c|}{Trans.} & \multicolumn{1}{c|}{Paper} & \multicolumn{1}{c|}{Mann.} & \multicolumn{1}{c|}{Obf.} & \multicolumn{1}{c|}{Imp.} & \multicolumn{1}{c|}{Cosm.} & \multicolumn{1}{c|}{FunnyEye} & \multicolumn{1}{c|}{P.Glasses} & \multicolumn{1}{c|}{P.Cut} & \multicolumn{1}{c|}{}                         \\ \hline
SVM+LBP \cite{7961798} & EER                       & 20.8                                         & 18.6                                        & 36.3                      & 21.4                          & 37.2                        & 7.5                        & 14.1                       & 51.2                      & 19.8                      & 16.1                       & 34.4                           & 33.0                               & 7.9                            & 51.2                                          \\
\rowcolor{gray!50} \cellcolor{gray!0} & ACER & 20.6 & 18.4 & 31.3 & 21.4 & 45.5 & 11.6 & 13.8 & 59.3 & 23.9 & 16.7 & 35.9 & 39.2 & 11.7 & 59.3 \\ \hline
Auxiliary \cite{Liu_2018_CVPR} & EER                    & 14.0                                         & 4.3                                         & 11.6                      & 12.4                          & 24.6                        & 7.8                        & 10.0                       & 72.3                      & 10.1                      & \textbf{9.4}               & 21.4                           & 18.6                               & 4.0                            & 72.3                                          \\ 
\rowcolor{gray!50} \cellcolor{gray!0} & ACER & 16.8 & 6.9  & 19.3 & 14.9 & 52.1 & 8.0  & 12.8 & 55.8 & 13.7 & 11.7 & 49.0 & 40.5 & 5.3  & 55.8 \\ \hline
DTN \cite{Liu_2019_CVPR} & EER                          & 10.0                                         & \textbf{2.1}                                & 14.4                      & 18.6                          & 26.5                        & 5.7                        & 9.6                        & 50.2                      & 10.1                      & 13.2                       & \textbf{19.8}                  & 20.5                               & 8.8                            & 50.2                                          \\ 
\rowcolor{gray!50} \cellcolor{gray!0} & ACER & 9.8  & 6.0  & 15.0 & 18.7 & 36.0 & 4.5  & 7.7  & 48.1 & 11.4 & 14.2 & 19.3 & 19.8 & 8.5  & 48.1 \\ \hline
CDC \cite{Yu_2020_CVPR} & EER                           & 9.2                                          & 5.6                                         & 4.2                       & 11.1                          & \textbf{19.3}               & 5.9                        & 5.0                        & 43.5                      & \textbf{0.0}              & 14.0                       & 23.3                           & 14.3                               & \textbf{0.0}                   & 43.5                                          \\ 
\rowcolor{gray!50} \cellcolor{gray!0} & ACER & 10.8 & 7.3  & 9.1  & 10.3 & 18.8 & 3.5  & 5.6  & 42.1 & 0.8  & 14.0 & 24.0 & 17.6 & 1.9  & 42.1 \\ \hline
LLIG \cite{deb2020look} & EER                         & \textbf{6.8}                                 & 11.2                                        & \textbf{2.8}              & \textbf{6.3}                  & 28.5                        & \textbf{0.4}               & \textbf{3.3}               & 17.8                      & 3.9                       & 11.7                       & 21.6                           & \textbf{13.5}                      & 3.6                            & 28.5                                          \\
\rowcolor{gray!50} \cellcolor{gray!0} & ACER & 7.4  & 19.5 & 3.2  & 7.7  & 33.3 & 5.2  & 3.3  & 22.5 & 5.9  & 11.7 & 21.7 & 14.1 & 6.4  & 33.3 \\ \hline
One-class & EER                         & 15.2                                         & 12.5                                        & 14.5                      & 25.8                          & 27.1                        & 23.9                       & 11.6                       & 23.0                      & 3.2                       & 23.1                       & 24.2                           & 26.7                               & 15.1                           & \textbf{27.1}                                 \\ 
\rowcolor{gray!50} \cellcolor{gray!0} & ACER & 13.2 & 10.9 & 11.1 & 23.4 & 30.9 & 18.1 & 9.3  & 18.0 & 4.8  & 19.7 & 16.1 & 24.0 & 11.2 & 30.9 \\ \hline
C-marmax & EER                      & 12.2                                         & 10.5                                        & 27.5                      & 8.2                           & 22.7                        & 13.4                       & 4.3                        & \textbf{17.0}             & 0.8                       & 11.3                       & 33.5                           & 33.2                               & 11.1                           & 33.5                                          \\ 
\rowcolor{gray!50} \cellcolor{gray!0} & ACER & 13.2 & 8.6  & 21.0 & 11.3 & 16.5 & 6.7  & 6.3  & 11.2 & 4.0  & 14.4 & 41.1 & 44.4 & 9.7  & 44.4 \\ \hline
Fusion & EER                        & 10.4                                         & 6.5                                         & 20.3                      & 10.9                          & 24.6                        & 3.0                        & 3.5                        & 23.0                      & 1.5                       & 12.8                       & 25.1                           & 27.8                               & 7.1                            & 27.8                                          \\ 
\rowcolor{gray!50} \cellcolor{gray!0} & ACER & 10.9 & 7.9  & 13.4 & 15.8 & 22.5 & 6.7  & 4.8  & 18.7 & 4.8  & 18.9 & 19.2 & 25.5 & 5.9  & 25.5 \\ \hline
\end{tabular}
\vspace{10pt}
\caption[Performance of the detector in few-shot learning scenarios on the SiW-M dataset]{Performance of the detector in few-shot learning scenarios on the SiW-M dataset.}
\label{few-shot}
\centering
\begin{tabular}{|l|l|r|r|r|r|r|r|r|r|r|r|r|r|r|r|}
\hline
\multirow{2}{*}{Method} & \multirow{2}{*}{Metric {[}\%{]}} & \multicolumn{1}{c|}{\multirow{2}{*}{Replay}} & \multicolumn{1}{c|}{\multirow{2}{*}{Print}} & \multicolumn{5}{c|}{Mask}                                                                                                                         & \multicolumn{3}{c|}{Makeup}                                                        & \multicolumn{3}{c|}{Partial}                                                                         & \multicolumn{1}{c|}{\multirow{2}{*}{MPA}} \\
                              & & \multicolumn{1}{c|}{}                        & \multicolumn{1}{c|}{}                       & \multicolumn{1}{c|}{Half} & \multicolumn{1}{c|}{Silicone} & \multicolumn{1}{c|}{Trans.} & \multicolumn{1}{c|}{Paper} & \multicolumn{1}{c|}{Mann.} & \multicolumn{1}{c|}{Obf.} & \multicolumn{1}{c|}{Imp.} & \multicolumn{1}{c|}{Cosm.} & \multicolumn{1}{c|}{FunnyEye} & \multicolumn{1}{c|}{P.Glasses} & \multicolumn{1}{c|}{P.Cut} & \multicolumn{1}{c|}{}                         \\ \hline
Zero shot~~~~~~~~~ & EER                     & 12.2                                         & 10.5                                         & 27.5                     & 8.2                           & 22.7                        & 13.4                       & 4.3                        & 17.0                      & \textbf{0.8}              & 11.3                       & 33.5                           & 33.2                               & 11.1                            & 33.5                                          \\ 
\rowcolor{gray!50} \cellcolor{gray!0} & ACER & 13.2 & 8.6  & 21.0 & 11.3 & 16.5 & 6.7 & 6.3 & 11.2 & 4.0 & 14.4 & 41.1 & 44.4 & 9.7  & 44.4 \\
 \hline
One shot & EER                      & 14.8                                         & 10.5                                        & 22.6                      & 7.6                           & 18.3                        & 6.2                        & 4.3               & 15.9                      & \textbf{0.8}                       & 14.7                       & 30.2                           & 26.4                               & 13.3                           & 30.2                                          \\ 
\rowcolor{gray!50} \cellcolor{gray!0} & ACER & 15.5 & 11.7 & 24.8 & 12.9 & 18.0 & 6.2 & 4.8 & 18.2 & 2.5 & 19.0 & 42.7 & 46.0 & 10.5 & 46.0 \\ \hline
Five shot & EER                     & 16.4                                         & 10.0                                        & 6.0                       & 9.8                           & 15.4                        & \textbf{3.0}               & 11.0                       & 15.9                      & 0.9                       & 18.3                       & 17.0                           & 15.0                               & \textbf{6.1}                   & 18.3                                          \\ 
\rowcolor{gray!50} \cellcolor{gray!0} & ACER & 16.3 & 8.8  & 9.8  & 9.1  & 12.1 & 7.2 & 8.9 & 11.1 & 2.7 & 12.4 & 19.2 & 23.3 & 14.5 & 23.3 \\ \hline
Known & EER                           & \textbf{9.7}                                          & \textbf{5.6}                                         & \textbf{1.5}                       & \textbf{6.6}                  & \textbf{4.5}                         & \textbf{3.0}                        & \textbf{3.8}               & \textbf{8.0}                       & 3.1                       & \textbf{7.8}               & \textbf{5.7}                            & \textbf{7.7}                                & 6.5                            & \textbf{9.7}                                  \\ 
\rowcolor{gray!50} \cellcolor{gray!0} & ACER & 10.6 & 5.6  & 4.3  & 6.6  & 8.0  & 8.6 & 6.3 & 13.3 & 4.7 & 9.3  & 6.3  & 9.2  & 6.5  & 13.3 \\ \hline
\end{tabular}
\vspace{10pt}
\caption[Performance of the proposed detection methods for the protocol II task of OULU-NPU dataset]{\label{ouludata}\note{Table added.}\change{}{Performance of the proposed detection methods for the protocol II task of OULU-NPU dataset.}}
\label{oulu}
\centering
\begin{tabular}{|l|r|r|r|r|r|r|r|r|}
\hline
\multicolumn{1}{|l|}{Metric {[}\%{]}} & Gradient \cite{boulkenafet2017competition} & Auxiliary \cite{Liu_2018_CVPR} & DeepPixBiS \cite{george2019deep} & TSCNN-ResNet \cite{chen2019attention} & LLIG \cite{deb2020look} & One-class & C-marmax     & Fusion \\ \hline
EER                    & \textbf{0.9}      & -       & -        & 2.0          & -     & 26.6      & 2.4 & 3.3    \\ \hline
\rowcolor{gray!50} ACER  & 2.5      & 2.7       & 6.0        & 4.9          & 3.4     & 52.4      & 3.1 & 3.1    \\ \hline
\end{tabular}
\end{sidewaystable*}

\subsubsection{OULU-NPU dataset}
\begin{figure}[htbp]
\centering\includegraphics[width=0.75\columnwidth]{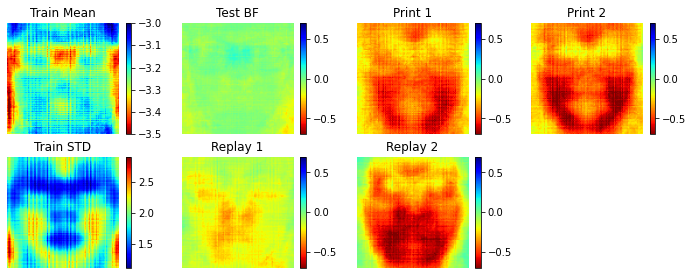}
\caption[Average and standard deviation of the log-likelihood matrices in the OULU-NPU dataset]{\label{oulufig}\note{Figure added.}\change{}{Average and standard deviation of the log-likelihood matrices over training data in the first column, along with the average log-likelihood matrices for test BF and each individual attack species in the OULU-NPU dataset in the following order: Printer 1, Printer 2, Replay 1, Replay 2.}}
\label{oulu_average}
\end{figure}

\change{}{In Table }\ref{ouludata}\change{}{, we report results of proposed framework as well as previously presented strategies in the literature on OULU-NPU dataset }\cite{7961798}\change{}{. Fig. }\ref{oulufig}\change{}{ shows the average and standard deviation of log-likelihood matrices for training data along with the average matrices for test data. The OULU-NPU face presentation attack detection database is composed of 4950 real access and attack videos. The videos were captured utilizing the front cameras of six mobile devices in three sessions with different background scenes and illumination conditions. There are two attacks, i.e., print and video-replay, which were generated via two printers and two display devices. In this study, we adopted OULU-NPU Protocol II because it presents the unknown attack detection scenario, namely, the effect of attack variation is assessed by introducing previously unseen print and video-replay attacks in the test set. We can observe in Table }\ref{ouludata}\change{}{ that the performance obtained using proposed framework is better than prior methods. For instance, the presented method with c-marmax achieved 2.4\% EER, whereas the scheme proposed in }\cite{george2019deep}\change{}{ obtained 6.0\% EER.} \change{}{It is also worth noticing that proposed system with one-class classifier did not perform well, but the proposed system with fusion scheme could avoid a major loss and attain notable accuracy.}

\subsubsection{Few-shot learning}
In Table \ref{few-shot}, the performance of the proposed method is presented on the task of few-shot learning when having one or five examples, and compared to unknown and known cases. \change{It can be seen that by observation of even one example from an unknown PAS, the performance of the system improves, and the MPA EER is reduced by $45\%$ to $18.3\%$ by observation of five examples.}{It can be seen that by observation of even one example from an unknown PAS, the performance of the system improves, and the MPA EER is reduced by $45\%$ from $33.5\%$ to $18.3\%$ by observation of five examples.} As such, the proposed system shows the capacity of significantly reducing the EER after the presentation of a few examples of a new PAS. It can also be seen that, specifically in the case of impersonation makeup attack, the observation of new samples does not reduce the EER. This can be explained by the fact that the EER is already low in the zero-shot case, and the proposed C-marmax loss does not reward further improvement in the EER of this attack as it does not improve the overall MPA EER.

\subsection{Detection Cost}
Due to the big size of the PixelCNN++ model, the extraction of each individual pixel log-likelihood matrix for each frame is the bottleneck and takes roughly $75$ milliseconds in our setup. Considering the average length of six seconds for the $24$ FPS videos in the dataset, processing each video takes $9$ seconds, corresponding to $\times1.5$ real-time speed. This may account for a prohibitively high detection cost in certain applications such as smartphone-based detection or social media monitoring. However, according to Eq. \ref{detect_payoff} the proposed method can find applications where the cost of a missed detection is high, such as border control and authenticity verification in journalism.

\section{Deepfake Detection}
\label{deepfake}

\begin{figure}[htbp]
\centering\includegraphics[width=0.75\columnwidth]{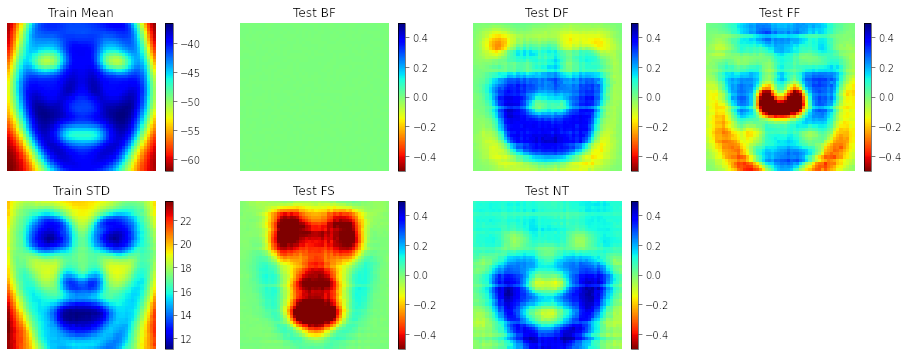}
\caption[Average and standard deviation of the log-likelihood matrices in the FaceForencisc++ dataset]{Average and standard deviation of the log-likelihood matrices over training data in the first column, along with the average log-likelihood matrices for test BF and each individual attack species in the FaceForencisc++ dataset in the following order: Deepfakes, Face2Face, FaceSwap, NeuralTextures.}
\label{df_average}
\end{figure}

Fig. \ref{df_average} shows the average and standard deviation of log-likelihood matrices for training data along with the average matrices for test data. It can be seen that most variations in the data are from the background, forehead, and cheeks, while the eye and mouth regions had little variability with a low log-likelihood average. The BF test data average matches that of training BF data. However, there are distinct patterns corresponding to each attack species. In the case of Deepfakes and NeuralTextures, there is a high log-likelihood region on the lower half of the face, corresponding to the possible over-smoothness of the texture. In the case of Deepfakes, there is a low-likelihood region around the eyebrows and the chin line which corresponds to the locations where the artifacts that are the characteristic of Deepfakes often occur. For the Face2Face technique, the pattern corresponds to points with low log-likelihood around the nose and chin line, while for the FaceSwap technique, the pattern corresponds to the eyes, nose, and mouth regions.

\begin{table}[htbp]
\begin{minipage}{\columnwidth}
\caption[Performance of the proposed detection methods for the task of known attack detection on Deepfake detection task on the FaceForensics++ dataset]{Performance of the proposed detection methods for the task of known attack detection on Deepfake detection task on the FaceForensics++ dataset.}
\label{known_df}
\centering
\begin{tabular}{|l|l|r|r|r|r|r|}
\hline
Method & Metric {[}\%{]} & DeepFake   & Face2Face   & FaceSwap   & NTexture   & MPA  \\ \hline
One-class   & EER & 6.43 & 8.21 & 2.14 & 2.14 & 8.21 \\ 
\rowcolor{gray!50} \cellcolor{gray!0} & ACER & 5.00 & 8.21 & 3.21 & 3.21 & 8.21 \\ \hline
C-marmax    & EER & 0.00 & 0.71 & 0.00 & 0.36 & 0.71 \\
\rowcolor{gray!50} \cellcolor{gray!0} & ACER & 2.50 & 2.50 & 2.50 & 2.50 & 2.50 \\ \hline
Fusion      & EER & 0.71 & 0.36 & 0.00 & 0.71 & 0.71 \\
\rowcolor{gray!50} \cellcolor{gray!0} & ACER & 2.50 & 2.50 & 2.50 & 2.50 & 2.50 \\ \hline
\end{tabular}
\end{minipage}
\end{table}

Table \ref{known_df} shows the performance of the one-class detector and the proposed discriminative detector as well as their fusion. It can be seen that the one-class detector managed to achieve acceptable MPA EER of $8.21\%$ while the discriminative detector achieved near-perfect video level detection. The Fusion did not degrade the performance of the discriminative detector significantly. It is important to mention that known attack detection on the raw subset of the dataset is a solved problem with near-perfect frame-level detection rates reported in the baseline \cite{Rossler_2019_ICCV}. 

Table \ref{unknown_df} reports the detection performance on the LOO unknown attack detection scenario. The low EERs of the discriminative detector shows that there are mutually discriminative features across the known and unknown attacks, especially for Face2Face and NeuralTexture methods. However, the face swap method shows less similarity to other methods and this results in a increase in the EER of the discriminative detector compared to the one-class one. Furthermore, the fusion mechanism managed to lower the MPA EER significantly, and an MPA EER of $2.5\%$ is achieved for the unknown attack detection. Due to the easiness of spotting digital manipulation traces in raw videos, the overall performances in terms of MPA are much lower than for PAD experiments.

\begin{table}[htbp]
\begin{minipage}{\columnwidth}
\caption[Performance of the proposed detection methods for the task of unknown attack detection on Deepfake detection task on the FaceForensics++ dataset]{Performance of the proposed detection methods for the task of unknown attack detection on Deepfake detection task on the FaceForensics++ dataset.}
\label{unknown_df}
\centering
\begin{tabular}{|l|l|r|r|r|r|r|}
\hline
Method & Metric {[}\%{]} & DeepFake   & Face2Face   & FaceSwap   & NTexture   & MPA  \\ \hline
One-class   & EER  & 6.43 & 8.21 & 2.14 & 2.14 & 8.21 \\
\rowcolor{gray!50} \cellcolor{gray!0} & ACER & 5.00 & 8.21 & 3.21 & 3.21 & 8.21 \\ \hline
C-marmax    & EER  & 5.36 & 1.07 & 5.71 & 1.79 & 5.71 \\
\rowcolor{gray!50} \cellcolor{gray!0} & ACER & 5.00 & 2.86 & 5.71 & 2.86 & 5.71 \\ \hline
Fusion      & EER  & 2.50 & 1.43 & 2.50 & 1.43 & 2.50 \\
\rowcolor{gray!50} \cellcolor{gray!0} & ACER & 3.57 & 2.50 & 4.29 & 2.50 & 4.29 \\ \hline
\end{tabular}
\end{minipage}
\end{table}

\section{Conclusion}
\label{conclusion}
\change{The choice of the attack by a rational attacker can have a significant negative impact on the performance of the detection systems in real-life scenarios. In response, after relying on game theory to build a theoretic basis and formulating the interactions between the attacker and the defender, a new detection method is proposed to optimize the performance against attacks from such attackers. Experiments on the tasks of presentation attack detection and Deepfake detection show the effectiveness of the proposed method in improving the detection rate on most powerful attacks both in known attack cases and when the detector faces unknown attacks. Furthermore, the proposed feature set is capable of enabling few-shot learning and explainability at pixel-level.
The proposed method shows generalizability across widely different types of attacks ranging from Deepfakes and replay attacks to 3D masks and makeup attacks and is able to show where the artifacts commonly occur for each specific attack species. Furthermore, the unsupervised anomaly detection method used is able to produce representations that cluster attacks from the same species together and separate BF samples from attacks in an unsupervised manner with limited training data in unconstrained recording conditions. }{The choice of the attack by a rational attacker can have a significant negative impact on the performance of the detection systems in real-life scenarios. In response, after relying on game theory to build a theoretic basis and formulating the interactions between the attacker and the defender, a new detection method is proposed to optimize the performance against attacks from such attackers. Experiments on the tasks of presentation attack detection and Deepfake detection show effectiveness of proposed method in improving detection rate on most powerful attacks both in known attack cases and when the detector faces unknown attacks. Furthermore, the proposed feature set is capable of enabling few-shot learning and explainability at pixel-level.
The proposed method shows generalizability across widely different types of attacks ranging from Deepfakes and replay attacks to 3D masks and makeup attacks and is able to show where the artifacts commonly occur for each specific attack species. Also, unsupervised anomaly detection method used is able to produce representations that cluster attacks from the same species together and separate BF samples from attacks in an unsupervised manner with limited training data in unconstrained recording conditions. }

However, this method has two specific short-comings. First, the extraction of the anomaly representations is computationally expensive and thus the system cannot be deployed in applications where processing an input video should be done faster than in real-time such as automated content monitoring on social media. Secondly, despite the proposed method outperforming the state-of-the-art in the task of presentation attack detection, its expected $27.8\%$ performance against the most powerful unknown attack is still far from acceptable for real-life applications, showing the need for further research in this direction. However, the availability of more training data from a more diverse set of attacks may alleviate this limitation.

\section{Acknowledgment}
This research work was funded by the Department of Information Security and Communication Technology at the Norwegian University of Science and Technology.

\bibliographystyle{unsrt}  
\bibliography{references}

\begin{thebibliography}{10}

\bibitem{Bohme2013}
Rainer B{\"o}hme and Matthias Kirchner.
\newblock {\em Counter-Forensics: Attacking Image Forensics}, pages 327--366.
\newblock Springer New York, New York, NY, 2013.

\bibitem{6515027}
M.~C. {Stamm}, M.~{Wu}, and K.~J.~R. {Liu}.
\newblock Information forensics: An overview of the first decade.
\newblock {\em IEEE Access}, 1:167--200, 2013.

\bibitem{4761645}
C.~{Chen}, Y.~Q. {Shi}, and W.~{Su}.
\newblock A machine learning based scheme for double jpeg compression
  detection.
\newblock In {\em 2008 19th ICPR}, pages 1--4, 2008.

\bibitem{10.1145/2037252.2037256}
Xunyu Pan, Xing Zhang, and Siwei Lyu.
\newblock Exposing image forgery with blind noise estimation.
\newblock MM\&amp;Sec '11, page 15–20, New York, NY, USA, 2011. ACM.

\bibitem{7004806}
A.~{De Rosa}, M.~{Fontani}, M.~{Massai}, A.~{Piva}, and M.~{Barni}.
\newblock Second-order statistics analysis to cope with contrast enhancement
  counter-forensics.
\newblock {\em IEEE Signal Processing Letters}, 22(8):1132--1136, 2015.

\bibitem{7090993}
F.~{Zhang}, P.~P.~K. {Chan}, B.~{Biggio}, D.~S. {Yeung}, and F.~{Roli}.
\newblock Adversarial feature selection against evasion attacks.
\newblock {\em IEEE Transactions on Cybernetics}, 46(3):766--777, 2016.

\bibitem{10.1007/978-3-319-20248-8_15}
Battista Biggio and et~al.
\newblock One-and-a-half-class multiple classifier systems for secure learning
  against evasion attacks at test time.
\newblock In {\em Multiple Classifier Systems}, pages 168--180, Cham, 2015.
  Springer.

\bibitem{8653427}
Z.~{Chen} and et~al.
\newblock Secure detection of image manipulation by means of random feature
  selection.
\newblock {\em IEEE Trans. on Info. For. and Sec.}, 14(9):2454--2469, 2019.

\bibitem{7823902}
M.~{Barni}, Z.~{Chen}, and B.~{Tondi}.
\newblock Adversary-aware, data-driven detection of double jpeg compression:
  How to make counter-forensics harder.
\newblock In {\em 2016 IEEE Int'l Workshop on Information Forensics and
  Security (WIFS)}, pages 1--6, 2016.

\bibitem{8081213}
M.~{Barni}, E.~{Nowroozi}, and B.~{Tondi}.
\newblock Higher-order, adversary-aware, double jpeg-detection via selected
  training on attacked samples.
\newblock In {\em 2017 25th European Signal Processing Conference (EUSIPCO)},
  pages 281--285, 2017.

\bibitem{8553305}
M.~{Barni}, M.~C. {Stamm}, and B.~{Tondi}.
\newblock Adversarial multimedia forensics: Overview and challenges ahead.
\newblock In {\em EUSIPCO}, pages 962--966, 2018.

\bibitem{6288237}
M.~C. {Stamm}, W.~S. {Lin}, and K.~J.~R. {Liu}.
\newblock Forensics vs. anti-forensics: A decision and game theoretic
  framework.
\newblock In {\em 2012 IEEE International Conference on Acoustics, Speech and
  Signal Processing (ICASSP)}, pages 1749--1752, 2012.

\bibitem{6400246}
M.~{Barni} and B.~{Tondi}.
\newblock The source identification game: An information-theoretic perspective.
\newblock {\em IEEE TIFS.}, 8(3):450--463, 2013.

\bibitem{4409068}
G.~{Pan}, L.~{Sun}, Z.~{Wu}, and S.~{Lao}.
\newblock Eyeblink-based anti-spoofing in face recognition from a generic
  webcamera.
\newblock In {\em 2007 IEEE 11th International Conference on Computer Vision},
  pages 1--8, 2007.

\bibitem{4291551}
K.~{Kollreider}, H.~{Fronthaler}, M.~I. {Faraj}, and J.~{Bigun}.
\newblock Real-time face detection and motion analysis with application in
  “liveness” assessment.
\newblock {\em IEEE Trans. on Info. Forensics and Security}, 2(3):548--558,
  2007.

\bibitem{10.1007/978-3-319-46654-5_67}
Keyurkumar Patel, Hu~Han, and Anil~K. Jain.
\newblock Cross-database face antispoofing with robust feature representation.
\newblock In Zhisheng You, Jie Zhou, Yunhong Wang, Zhenan Sun, Shiguang Shan,
  Weishi Zheng, Jianjiang Feng, and Qijun Zhao, editors, {\em Biometric
  Recognition}, pages 611--619, Cham, 2016. Springer International Publishing.

\bibitem{6612955}
J.~{Yang}, Z.~{Lei}, S.~{Liao}, and S.~Z. {Li}.
\newblock Face liveness detection with component dependent descriptor.
\newblock In {\em Int'l Conf. on Biometrics (ICB)}, pages 1--6, 2013.

\bibitem{10.1007/978-3-642-15567-3_37}
Xiaoyang Tan, Yi~Li, Jun Liu, and Lin Jiang.
\newblock Face liveness detection from a single image with sparse low rank
  bilinear discriminative model.
\newblock In Kostas Daniilidis, Petros Maragos, and Nikos Paragios, editors,
  {\em ECCV 2010}, pages 504--517, 2010.

\bibitem{7487030}
K.~{Patel}, H.~{Han}, and A.~K. {Jain}.
\newblock Secure face unlock: Spoof detection on smartphones.
\newblock {\em IEEE Trans. on Info. Forensics and Sec.}, 11(10):2268--2283,
  2016.

\bibitem{7748511}
Z.~{Boulkenafet}, J.~{Komulainen}, and A.~{Hadid}.
\newblock Face antispoofing using speeded-up robust features and fisher vector
  encoding.
\newblock {\em IEEE Signal Processing Letters}, 24(2):141--145, 2017.

\bibitem{7031384}
D.~{Wen}, H.~{Han}, and A.~K. {Jain}.
\newblock Face spoof detection with image distortion analysis.
\newblock {\em IEEE TIFS}, 10(4):746--761, 2015.

\bibitem{10.1007/978-3-642-37410-4_11}
de~Freitas Pereira~et al.
\newblock Lbp - top based countermeasure against face spoofing attacks.
\newblock In {\em ACCV Workshops}, pages 121--132, 2013.

\bibitem{Yu_2020_CVPR}
Zitong Yu, Chenxu Zhao, Zezheng Wang, Yunxiao Qin, Zhuo Su, Xiaobai Li, Feng
  Zhou, and Guoying Zhao.
\newblock Searching central difference convolutional networks for face
  anti-spoofing.
\newblock In {\em IEEE/CVF Conference on Computer Vision and Pattern
  Recognition}, June 2020.

\bibitem{Li2020compactnet}
Lei Li and et~al.
\newblock Compactnet: learning a compact space for face presentation attack
  detection.
\newblock {\em Neurocomputing}, 409:191 -- 207, 2020.

\bibitem{israel2019style}
Israel A.~Laurensi R., Luciana~T. Menon, Manoel Camillo O.~Penna N.,
  Alessandro~L. Koerich, and Alceu S. Britto~Jr au2.
\newblock Style transfer applied to face liveness detection with user-centered
  models, 2019.

\bibitem{WANG2017332}
Yan Wang, Fudong Nian, Teng Li, Zhijun Meng, and Kongqiao Wang.
\newblock Robust face anti-spoofing with depth information.
\newblock {\em J. of Vis. Comm. and Im. Rep.}, 49:332 -- 337, 2017.

\bibitem{5771438}
Z.~{Zhang}, D.~{Yi}, Z.~{Lei}, and S.~Z. {Li}.
\newblock Face liveness detection by learning multispectral reflectance
  distributions.
\newblock In {\em Face and Gesture}, pages 436--441, 2011.

\bibitem{5584864}
G.~{Chetty}.
\newblock Biometric liveness checking using multimodal fuzzy fusion.
\newblock In {\em International Conference on Fuzzy Systems}, pages 1--8, 2010.

\bibitem{6671991}
J.~{Galbally}, S.~{Marcel}, and J.~{Fierrez}.
\newblock Image quality assessment for fake biometric detection: Application to
  iris, fingerprint, and face recognition.
\newblock {\em IEEE Transactions on Image Processing}, 23(2):710--724, 2014.

\bibitem{7821027}
H.~{Li}, S.~{Wang}, and A.~C. {Kot}.
\newblock Face spoofing detection with image quality regression.
\newblock In {\em Int'l Conf. on Image Pro. Theory, Tools and App.}, pages
  1--6, 2016.

\bibitem{Liu_2018_CVPR}
Yaojie Liu, Amin Jourabloo, and Xiaoming Liu.
\newblock Learning deep models for face anti-spoofing: Binary or auxiliary
  supervision.
\newblock In {\em IEEE CVPR}, June 2018.

\bibitem{deb2020look}
Debayan Deb and Anil~K. Jain.
\newblock Look locally infer globally: A generalizable face anti-spoofing
  approach, 2020.

\bibitem{ramachandra2017presentation}
Raghavendra Ramachandra and Christoph Busch.
\newblock Presentation attack detection methods for face recognition systems: A
  comprehensive survey.
\newblock {\em Comp. Suv.}, 50(1), March 2017.

\bibitem{Bhattacharjee2019recent}
Sushil Bhattacharjee, Amir Mohammadi, Andr{\'e} Anjos, and S{\'e}bastien
  Marcel.
\newblock {\em Recent Advances in Face Presentation Attack Detection}, pages
  207--228.
\newblock Springer, 2019.

\bibitem{7984788}
S.~R. {Arashloo}, J.~{Kittler}, and W.~{Christmas}.
\newblock An anomaly detection approach to face spoofing detection: A new
  formulation and evaluation protocol.
\newblock {\em IEEE Access}, 5:13868--13882, 2017.

\bibitem{8411206}
O.~{Nikisins}, A.~{Mohammadi}, A.~{Anjos}, and S.~{Marcel}.
\newblock On effectiveness of anomaly detection approaches against unseen
  presentation attacks in face anti-spoofing.
\newblock In {\em International Conference on Biometrics}, pages 75--81, 2018.

\bibitem{Perez-Cabo_2019_CVPR_Workshops}
Daniel Perez-Cabo, David Jimenez-Cabello, Artur Costa-Pazo, and Roberto~J.
  Lopez-Sastre.
\newblock Deep anomaly detection for generalized face anti-spoofing.
\newblock In {\em Proceedings of the IEEE/CVF CVPR Workshops}, June 2019.

\bibitem{Liu_2019_CVPR}
Yaojie Liu, Joel Stehouwer, Amin Jourabloo, and Xiaoming Liu.
\newblock Deep tree learning for zero-shot face anti-spoofing.
\newblock In {\em CVPR}, June 2019.

\bibitem{6333919}
D.~{Dang-Nguyen}, G.~{Boato}, and F.~G.~B. {De Natale}.
\newblock Discrimination between computer generated and natural human faces
  based on asymmetry information.
\newblock In {\em European Sig. Proc. Conf. (EUSIPCO)}, pages 1234--1238, 2012.

\bibitem{7097661}
D.~{Dang-Nguyen}, G.~{Boato}, and F.~G.~B. {De Natale}.
\newblock 3d-model-based video analysis for computer generated faces
  identification.
\newblock {\em IEEE Trans. on Info. Forensics and Security}, 10(8):1752--1763,
  2015.

\bibitem{7025049}
V.~{Conotter}, E.~{Bodnari}, G.~{Boato}, and H.~{Farid}.
\newblock Physiologically-based detection of computer generated faces in video.
\newblock In {\em ICIP}, pages 248--252, 2014.

\bibitem{8638330}
F.~{Matern}, C.~{Riess}, and M.~{Stamminger}.
\newblock Exploiting visual artifacts to expose deepfakes and face
  manipulations.
\newblock In {\em 2019 IEEE Winter Applications of Computer Vision Workshops
  (WACVW)}, pages 83--92, 2019.

\bibitem{8630787}
Y.~{Li}, M.~{Chang}, and S.~{Lyu}.
\newblock In ictu oculi: Exposing ai created fake videos by detecting eye
  blinking.
\newblock In {\em 2018 IEEE International Workshop on Information Forensics and
  Security (WIFS)}, pages 1--7, 2018.

\bibitem{9141516}
U.~A. {Ciftci}, I.~{Demir}, and L.~{Yin}.
\newblock Fakecatcher: Detection of synthetic portrait videos using biological
  signals.
\newblock {\em IEEE TPAMI}, pages 1--1, 2020.

\bibitem{Li_2019_CVPR_Workshops}
Yuezun Li and Siwei Lyu.
\newblock Exposing deepfake videos by detecting face warping artifacts.
\newblock In {\em IEEE/CVF CVPR Workshops}, June 2019.

\bibitem{10.1145/3335203.3335724}
Xin Yang, Yuezun Li, Honggang Qi, and Siwei Lyu.
\newblock Exposing gan-synthesized faces using landmark locations.
\newblock In {\em ACM Workshop on Information Hiding and Multimedia Security},
  IH\&amp;MMSec'19, page 113–118, 2019.

\bibitem{8683164}
X.~{Yang}, Y.~{Li}, and S.~{Lyu}.
\newblock Exposing deep fakes using inconsistent head poses.
\newblock In {\em IEEE ICASSP}, pages 8261--8265, 2019.

\bibitem{8630761}
D.~{Afchar}, V.~{Nozick}, J.~{Yamagishi}, and I.~{Echizen}.
\newblock Mesonet: a compact facial video forgery detection network.
\newblock In {\em 2018 IEEE International Workshop on Information Forensics and
  Security (WIFS)}, pages 1--7, 2018.

\bibitem{DBLP:journals/corr/abs-1808-07276}
Haodong Li, Bin Li, Shunquan Tan, and Jiwu Huang.
\newblock Detection of deep network generated images using disparities in color
  components.
\newblock {\em CoRR}, abs/1808.07276, 2018.

\bibitem{8803661}
S.~{McCloskey} and M.~{Albright}.
\newblock Detecting gan-generated imagery using saturation cues.
\newblock In {\em IEEE Int'l Conf. on Image Processing}, pages 4584--4588,
  2019.

\bibitem{8695364}
F.~{Marra}, D.~{Gragnaniello}, L.~{Verdoliva}, and G.~{Poggi}.
\newblock Do gans leave artificial fingerprints?
\newblock In {\em IEEE Conference on MIPR}, pages 506--511, 2019.

\bibitem{Rossler_2019_ICCV}
Andreas Rossler, Davide Cozzolino, Luisa Verdoliva, Christian Riess, Justus
  Thies, and Matthias Niessner.
\newblock Faceforensics++: Learning to detect manipulated facial images.
\newblock In {\em IEEE/CVF International Conference on Computer Vision (ICCV)},
  October 2019.

\bibitem{Dang_2020_CVPR}
Hao Dang, Feng Liu, Joel Stehouwer, Xiaoming Liu, and Anil~K. Jain.
\newblock On the detection of digital face manipulation.
\newblock In {\em IEEE/CVF CVPR}, June 2020.

\bibitem{8682602}
H.~H. {Nguyen}, J.~{Yamagishi}, and I.~{Echizen}.
\newblock Capsule-forensics: Using capsule networks to detect forged images and
  videos.
\newblock In {\em IEEE Int'l Conf. on Acoustics, Speech and Signal Processing},
  pages 2307--2311, 2019.

\bibitem{8639163}
D.~{Güera} and E.~J. {Delp}.
\newblock Deepfake video detection using recurrent neural networks.
\newblock In {\em IEEE AVSS}, pages 1--6, 2018.

\bibitem{Sabir_2019_CVPR_Workshops}
Ekraam Sabir, Jiaxin Cheng, Ayush Jaiswal, Wael AbdAlmageed, Iacopo Masi, and
  Prem Natarajan.
\newblock Recurrent convolutional strategies for face manipulation detection in
  videos.
\newblock In {\em IEEE/CVF CVPR Workshops}, June 2019.

\bibitem{Amerini_2019_ICCV}
Irene Amerini, Leonardo Galteri, Roberto Caldelli, and Alberto Del~Bimbo.
\newblock Deepfake video detection through optical flow based cnn.
\newblock In {\em Proceedings of the IEEE/CVF International Conference on
  Computer Vision (ICCV) Workshops}, Oct 2019.

\bibitem{8553251}
A.~{Khodabakhsh}, R.~{Ramachandra}, K.~{Raja}, P.~{Wasnik}, and C.~{Busch}.
\newblock Fake face detection methods: Can they be generalized?
\newblock In {\em 2018 International Conference of the Biometrics Special
  Interest Group}, pages 1--6, 2018.

\bibitem{DBLP:journals/corr/abs-1812-02510}
Davide Cozzolino, Justus Thies, Andreas R{\"{o}}ssler, Christian Riess,
  Matthias Nie{\ss}ner, and Luisa Verdoliva.
\newblock Forensictransfer: Weakly-supervised domain adaptation for forgery
  detection.
\newblock {\em CoRR}, abs/1812.02510, 2018.

\bibitem{du2020generalizable}
Mengnan Du, Shiva Pentyala, Yuening Li, and Xia Hu.
\newblock Towards generalizable deepfake detection with locality-aware
  autoencoder, 2020.

\bibitem{9035099}
F.~{Marra}, C.~{Saltori}, G.~{Boato}, and L.~{Verdoliva}.
\newblock Incremental learning for the detection and classification of
  gan-generated images.
\newblock In {\em 2019 IEEE International Workshop on Information Forensics and
  Security}, pages 1--6, 2019.

\bibitem{10.1007/978-3-030-31456-9_15}
Xinsheng Xuan, Bo~Peng, Wei Wang, and Jing Dong.
\newblock On the generalization of gan image forensics.
\newblock In Zhenan Sun, Ran He, Jianjiang Feng, Shiguang Shan, and Zhenhua
  Guo, editors, {\em Biometric Recognition}, pages 134--141, Cham, 2019.
  Springer International Publishing.

\bibitem{wang2020cnngenerated}
Sheng-Yu Wang, Oliver Wang, Richard Zhang, Andrew Owens, and Alexei~A. Efros.
\newblock Cnn-generated images are surprisingly easy to spot... for now, 2020.

\bibitem{fernando2019exploiting}
Tharindu Fernando, Clinton Fookes, Simon Denman, and Sridha Sridharan.
\newblock Exploiting human social cognition for the detection of fake and
  fraudulent faces via memory networks, 2019.

\bibitem{Li_2020_CVPR}
Lingzhi Li, Jianmin Bao, Ting Zhang, Hao Yang, Dong Chen, Fang Wen, and Baining
  Guo.
\newblock Face x-ray for more general face forgery detection.
\newblock In {\em CVPR}, June 2020.

\bibitem{Verdoliva_2019_CVPR_Workshops}
Davide Cozzolino Giovanni Poggi Luisa~Verdoliva.
\newblock Extracting camera-based fingerprints for video forensics.
\newblock In {\em CVPR Workshops}, June 2019.

\bibitem{akhtar2020utility}
Z.~{Akhtar}, M.~R. {Mouree}, and D.~{Dasgupta}.
\newblock Utility of deep learning features for facial attributes manipulation
  detection.
\newblock In {\em IEEE Int'l Conf. on Humanized Computing and Communication
  with Artificial Intelligence}, pages 55--60, 2020.

\bibitem{tolosana2020deepfakes}
Ruben Tolosana, Ruben Vera-Rodriguez, Julian Fierrez, Aythami Morales, and
  Javier Ortega-Garcia.
\newblock Deepfakes and beyond: A survey of face manipulation and fake
  detection, 2020.

\bibitem{perera2020generative}
Pramuditha Perera, Vlad~I Morariu, Rajiv Jain, Varun Manjunatha, Curtis
  Wigington, Vicente Ordonez, and Vishal~M Patel.
\newblock Generative-discriminative feature representations for open-set
  recognition.
\newblock In {\em IEEE/CVF Conference on Computer Vision and Pattern
  Recognition}, pages 11814--11823, 2020.

\bibitem{neal2018open}
Lawrence Neal, Matthew Olson, Xiaoli Fern, Weng-Keen Wong, and Fuxin Li.
\newblock Open set learning with counterfactual images.
\newblock In {\em ECCV}, pages 613--628, 2018.

\bibitem{ditria2020opengan}
Luke Ditria, Benjamin~J Meyer, and Tom Drummond.
\newblock Opengan: Open set generative adversarial networks.
\newblock In {\em Asian Conference on Computer Vision}, 2020.

\bibitem{goodfellow2014generative}
Ian Goodfellow, Jean Pouget-Abadie, Mehdi Mirza, Bing Xu, David Warde-Farley,
  Sherjil Ozair, Aaron Courville, and Yoshua Bengio.
\newblock Generative adversarial nets.
\newblock In Z.~Ghahramani et~al., editor, {\em Adv. in Neural Info. Proc.
  Sys.}, volume~27, pages 2672--2680, 2014.

\bibitem{DBLP:journals/corr/OordKK16}
A{\"{a}}ron van~den Oord, Nal Kalchbrenner, and Koray Kavukcuoglu.
\newblock Pixel recurrent neural networks.
\newblock {\em CoRR}, abs/1601.06759, 2016.

\bibitem{schroff2015facenet}
Florian Schroff, Dmitry Kalenichenko, and James Philbin.
\newblock Facenet: A unified embedding for face recognition and clustering.
\newblock In {\em CVPR}, June 2015.

\bibitem{Thies_2016_CVPR}
Justus Thies, Michael Zollhofer, Marc Stamminger, Christian Theobalt, and
  Matthias Niessner.
\newblock Face2face: Real-time face capture and reenactment of rgb videos.
\newblock In {\em IEEE Conference on Computer Vision and Pattern Recognition},
  June 2016.

\bibitem{10.1145/3306346.3323035}
Justus Thies, Michael Zollh\"{o}fer, and Matthias Nie\ss{}ner.
\newblock Deferred neural rendering: Image synthesis using neural textures.
\newblock {\em ACM Trans. Graph.}, 38(4), July 2019.

\bibitem{king2009dlib}
Davis~E King.
\newblock Dlib-ml: A machine learning toolkit.
\newblock {\em Journal of Machine Learning Research}, 10(Jul):1755--1758, 2009.

\bibitem{DBLP:journals/corr/SalimansKCK17}
Tim Salimans, Andrej Karpathy, Xi~Chen, and Diederik~P. Kingma.
\newblock Pixelcnn++: Improving the pixelcnn with discretized logistic mixture
  likelihood and other modifications.
\newblock {\em CoRR}, abs/1701.05517, 2017.

\bibitem{10.5555/3045390.3045624}
Wenling Shang, Kihyuk Sohn, Diogo Almeida, and Honglak Lee.
\newblock Understanding and improving convolutional neural networks via
  concatenated rectified linear units.
\newblock In {\em Int'l Conf. on Machine Learning}, ICML'16, page 2217–2225,
  2016.

\bibitem{DBLP:journals/corr/KingmaB14}
Diederik~P. Kingma and Jimmy Ba.
\newblock Adam: {A} method for stochastic optimization.
\newblock In Yoshua Bengio and Yann LeCun, editors, {\em 3rd Int'l Conf. on
  Learning Representations}, 2015.

\bibitem{10.5555/2968618.2968725}
Geoffrey Hinton and Sam Roweis.
\newblock Stochastic neighbor embedding.
\newblock In {\em Proceedings of the 15th International Conference on Neural
  Information Processing Systems}, NIPS'02, page 857–864, Cambridge, MA, USA,
  2002. MIT Press.

\bibitem{7961798}
Z.~{Boulkenafet}, J.~{Komulainen}, L.~{Li}, X.~{Feng}, and A.~{Hadid}.
\newblock Oulu-npu: A mobile face presentation attack database with real-world
  variations.
\newblock In {\em IEEE Int'l Conf. on Automatic Face Gesture Recog.}, pages
  612--618, 2017.

\bibitem{boulkenafet2017competition}
Z.~{Boulkenafet} and et~al.
\newblock A competition on generalized software-based face presentation attack
  detection in mobile scenarios.
\newblock In {\em 2017 IEEE International Joint Conference on Biometrics
  (IJCB)}, pages 688--696, 2017.

\bibitem{george2019deep}
A.~{George} and S.~{Marcel}.
\newblock Deep pixel-wise binary supervision for face presentation attack
  detection.
\newblock In {\em 2019 International Conference on Biometrics (ICB)}, pages
  1--8, 2019.

\bibitem{chen2019attention}
H.~{Chen}, G.~{Hu}, Z.~{Lei}, Y.~{Chen}, N.~M. {Robertson}, and S.~Z. {Li}.
\newblock Attention-based two-stream convolutional networks for face spoofing
  detection.
\newblock {\em IEEE Transactions on Information Forensics and Security},
  15:578--593, 2020.

\end{thebibliography}

\end{document}